%% file: main.tex
\documentclass{article}

\usepackage[nonatbib,preprint]{neurips_2022}

\usepackage[utf8]{inputenc} 
\usepackage[T1]{fontenc}    
\usepackage{hyperref}
\usepackage{url}            
\usepackage{booktabs}       
\usepackage{amsfonts}       
\usepackage{nicefrac}       
\usepackage{microtype}      
\usepackage{xcolor}         
\usepackage{float}
\usepackage{subfig}
\usepackage{wrapfig}
\usepackage{multirow}
\usepackage{bbding}
\usepackage{framed}
\usepackage{setspace}
\usepackage{pifont}
\usepackage{graphicx}
\usepackage{threeparttable}
\usepackage{times}
\usepackage{amsmath}
\usepackage{soul}
\usepackage{url}
\usepackage[utf8]{inputenc}
\usepackage{amsfonts}
\usepackage{graphicx}
\usepackage{amsmath, bm}
\usepackage{booktabs}
\usepackage{algorithm}
\usepackage{amsmath}
\usepackage{amssymb}
\usepackage{booktabs}
\usepackage[T1]{fontenc}
\usepackage{graphicx}
\usepackage{etoc}
\usepackage[most]{tcolorbox}
\usepackage{multirow}
\usepackage{amsmath}
\usepackage{float}
\usepackage{amsthm}
\usepackage{comment}
\usepackage{mathrsfs}
\usepackage{multicol}
\usepackage{appendix}
\usepackage{amssymb}
\usepackage{wrapfig}
\usepackage{bbm}
\usepackage{bbding}
\usepackage{framed}
\usepackage{setspace}
\usepackage{pifont}
\usepackage{algorithmic}
\usepackage{url}
\usepackage{color,xcolor}
\usepackage{xspace}
  
\newcommand{\fair}[0]{\textrm{fair}}
\newcommand{\greedy}[0]{G-fair-Prompting\xspace}
\newcommand{\topk}[0]{T-fair-Prompting\xspace}

\newcommand{\best}[1]{\bm{\textcolor{brown}{#1}}}
\newcommand{\view}[1]{\emph{\textcolor{brown}{#1}}}

\newcommand*{\affaddr}[1]{#1} 
\newcommand*{\affmark}[1][*]{\textsuperscript{#1}}
\newcommand*{\email}[1]{\small{\texttt{#1}}}
\newcommand*\samethanks[1][\value{footnote}]{\footnotemark[#1]}

\theoremstyle{plain}
\newtheorem{theorem}{Theorem}[section]

\title{Fairness-guided Few-shot Prompting for Large Language Models}

\author{%
Huan Ma\affmark[1,]\affmark[2]\thanks{The project was conducted during the internship in AI Lab, Tencent}, ~Changqing Zhang\affmark[2]\samethanks[2], ~Yatao Bian\affmark[1], Lemao Liu\affmark[1], Zhirui Zhang\affmark[1], \\ \textbf{Peilin Zhao\affmark[1], Shu Zhang\affmark[1], Huazhu Fu\affmark[3], Qinghua Hu\affmark[2], Bingzhe Wu\affmark[1]\thanks{Corresponding author}} \\ 
\affaddr{\affmark[1] AI Lab, Tencent, Shenzhen, China}\\
\affaddr{\affmark[2] College of Intelligence and Computing, Tianjin University, Tianjin, China}\\
\affaddr{\affmark[3] Institute of High Performance Computing, A*STAR, Singapore}
\\
\email{\affmark[2] \href{mailto:zhanchangqing@tju.edu.cn}{zhanchangqing@tju.edu.cn}};\; \email{\affmark[1] \href{mailto:wubingzheagent@gmail.com}{bingzhewu@tencent.com}}
}

\begin{document}

\maketitle

\begin{abstract}
Large language models have demonstrated surprising ability to perform in-context learning, i.e., these models can be directly applied to solve numerous downstream tasks by conditioning on a prompt constructed by a few input-output examples. However, prior research has shown that in-context learning can suffer from high instability due to variations in training examples, example order, and prompt formats. Therefore, the construction of an appropriate prompt is essential for improving the performance of in-context learning. 
In this paper,  we revisit this problem from the view of predictive bias. Specifically, we introduce a metric to evaluate the predictive bias of a fixed prompt against labels or a given attributes. Then we empirically show that prompts with higher bias always lead to unsatisfactory predictive quality. Based on this observation, we propose a novel search strategy based on the greedy search to identify the near-optimal prompt for improving the performance of in-context learning. 
We perform comprehensive experiments with state-of-the-art mainstream models such as GPT-3 on various downstream tasks. 
Our results indicate that our method can enhance the model's in-context learning performance in an effective and interpretable manner. Code is available at: \href{https://github.com/MaHuanAAA/g_fair_prompting}{https://github.com/MaHuanAAA.} 
\end{abstract}

\input{secs/intro.tex}

\input{secs/rela_wor.tex}

\input{secs/method.tex}
\input{secs/exp.tex}

\section{Conclusion}
In this paper, we revisit the sensitivity of large language model across prompts, and analyse the issue from a predictive bias perspective. Accordingly, we employ a "content-free" strategy as a metric termed as fairness to evaluate the predictive bias of a fixed prompt and show that model's performance is highly consistency with fairness. Then, we propose two strategy to search the most fair prompt in the original space. We conduct extensive experiments on current famous LLMs, and validate the effectiveness of the proposed strategy. Moreover, in addition to fairness adopted in this paper, there would be more metrics for prompt searching in the future for different scenarios. 

\bibliographystyle{unsrt}
\bibliography{reference}

\clearpage

\input{secs/appendix.tex}

\end{document}

%% file: secs/intro.tex
\section{Introduction}

Large language models (LLMs), such as GPT-3~\cite{gpt32020brown} and BLOOM~\cite{bloom2022}, have demonstrated remarkable ability in performing in-context learning (ICL) on downstream tasks. ICL refers to the process of conditioning an LLM to solve various downstream tasks using prompts constructed from a few demonstration input-output pairs \cite{petroni2019language} (i.e., few-shot prompting). Despite its impressive performance, prior research has shown that ICL suffers from high instability due to variations in the choice of in-context demonstrations, demonstration order, and prompt formats ~\cite{order2021lu,nie2022improving}. Therefore, constructing an appropriate prompt has been identified as a critical factor for improving the performance of ICL \cite{liu2023pre}.

Previous research studies this problem typically from two directions: (1) prompt tuning in the embedding space \cite{li2021prefix,liu2021p,hambardzumyan2021warp,qin2021learning,liu2021gpt} (2) prompt searching in the text space \cite{order2021lu,zhang2022automatic,gentile2022fast,diao2023active,liu2021makes,shi2023large}. The key idea of prompt tuning is to inject task-specific embedding into hidden layers and then tune these embeddings using gradient-based optimization \cite{liu2021p,liu2021makes}. However, these methods require to modify the original inference process of the model, which is impractical for the case of black-box LM services such as GPT3 and ChatGPT \cite{chatgpt}. Furthermore, prompt tuning introduces additional computational and storage costs, which is typically expensive for LLM. 
A more feasible and efficient way is to optimize prompting via searching approximate demonstration samples and ordering in the original text space \cite{order2021lu,liu2021makes}. Bunch of works are presented to constructs prompts from either "global" or "local" views. On the one hand, global-view based methods typically optimize the different elements of the prompt as a whole, with the aim of achieving superior performance. For example, one approach, as described in \cite{diao2023active}, constructs a search procedure that leverages the overall diversity of demonstrations. Another approach \cite{order2021lu} attempts to optimize the ordering of the entire set of demonstrations to achieve better performance. In contrast to the global view, local-view based methods optimize each individual demonstration by designing different heuristic selection criteria such as prior work KATE \cite{liu2021makes}.
These methods have achieved impressive improvements on a wide range of tasks. However, most of them still suffer from  the following limitations: 
(1) Most of current research mainly focuses on searching prompts along a single dimension, such as example selection or order. However, the overall influence of various dimensions on the performance remains unclear.
(2) These methods are typically based on heuristic criteria, and there is a gap between them and actual performance. A unified view that explains how these methods work is needed.
(3) More importantly, existing methods optimize prompts globally or locally, which may lead to suboptimal performance.

In this paper, we revisit this problem from the perspective of \emph{predictive bias}. We find a key insight that the quality of a given prompt depends on its inherent bias. Based on this insight, 
we propose a surrogate metric based on predictive bias for evaluating the quality of prompts. This metric allows us to evaluate a prompt in a single forward process without an additional development set. Specifically, 
we apply a given prompt to a "content-free" input and expect the model output an uniform predictive distribution (a content-free input contains no useful information). Therefore, we employ the uniformity of the predictive distribution to characterize the bias of a give prompt.  This shares a similar idea to the prior work which uses this metric to calibrate the model output \cite{calibrate2021zhao}. In contrast to this work which mainly focus on using this metric for calibration when the prompt is fixed, we further explore its usage  in automatically searching an approximate prompt. Moreover, through extensive experiments, we empirically validate the correlation between the inherent bias of a given prompt and its quality measured by the average task performance on a given test set (see Fig.~\ref{fig:allcandidates}). 

Moreover, this bias-based metric allows us to build prompting optimization techniques in a "local-to-global" manner. We present two novel strategies for efficiently searching high-quality prompts in a  bias-guided way: (1) \topk (2) \greedy . We focus on a general setting where a labeled  set with size $N$ is given. The goal of our strategies is to perform combinatorial optimization over this set to find near-optimal prompts (i.e., select demonstrations and their orders). Specifically, \topk  uses an intuitive way that first computes the bias of each single demonstration (i.e., one-shot prompting) and then select the top-k fair demonstrations to form the final prompts. This strategy can be efficiently done with a complexity of $O(N)$. Note that \topk is based on the assumption that the optimal prompt is usually constructed from demonstrations with the smallest individual bias. However, this may not hold true in real situations and often leads to sub-optimal solutions.
Therefore, we further introduce \greedy to improve the search quality. \greedy follows the normal procedure of the greedy search which  finds the optimal solution by making locally optimal choices at each step. At each step of the algorithm, the selected demonstration is the one which makes the updated prompts achieves the best fairness score.This strategy trades off the quality of the search with the worst-case time complexity. By accepting a higher worst-case time complexity of $O(N^2)$, the search quality is significantly improved. Note that \greedy works from a local to global perspective, wherein bias of individual samples are considered in the early stages while the later stage focus on the reduction of global predictive bias.

To evaluate the effectiveness of our strategies, we conduct extensive experiments with current mainstream models, such as GPT-3~\cite{gpt32020brown}, on various downstream tasks. Our results indicate that our method can significantly enhance the model's in-context learning performance in an effective and interpretable manner. The overall contribution is summarized as follows:
\begin{itemize}
    \item We introduce to use the predictive bias to assess the quality of a given prompt in an efficient and development set independent way and the empirical effectiveness of this metric is comprehensively validated.
    \item Based on the above idea, we propose two efficient and effective strategies, namely, \topk and \greedy to optimize the prompts.
    \item The effectiveness of these two strategies are validated on various LLMs ranging from GPT-series models to LMaMA family~\cite{touvron2023llama} released by Meta recently. Consistent relative improvements of over $10\%$ have been observed over different downstream tasks in contrast to SOTA methods.
\end{itemize}
\noindent \textbf{Relation to Calibration-before-use:}
Our paper shares a similar metric with cal-before-use~\cite{calibrate2021zhao} to asses the predictive bias of a given prompt. However, the prior approach aims to use this metric to calibrate the output, which can be still easily affected by the quality of the used prompt (more results can be found in Table~\ref{tab:calibrated}).  In contrast, our research aims to find a near-optimal prompt on the original space to improve the model's performance, without requiring any post-adjustment to the output of the model. Moreover, we have firstly empirically validated the connection between predictive bias and the final task performance as shown in Fig.~\ref{fig:allcandidates}, which has not been studied in \cite{calibrate2021zhao}.
Through experiments, we have discovered that, even without calibration, the prompt selected by our method can outperform a randomly selected prompt with calibration.

%% file: secs/rela_wor.tex
\section{Related Work}

\textbf{In-context Learning}\quad
 Previous research, as cited in \cite{gpt32020brown,gpt22018}, has demonstrated that Large Language Models can complete tasks with zero- or few-shot learning using in-context learning. LLMs perform well with an appropriate prompt. However, recent works~\cite{order2021lu,calibrate2021zhao} have shown that the performance of LLMs  is affected by the prompt used. Therefore, determining the optimal prompt is a crucial and fundamental research area.

\begin{figure}[t]
    \centering
    \subfloat[Selection]{
    \centering
    \includegraphics[width=0.24\linewidth]{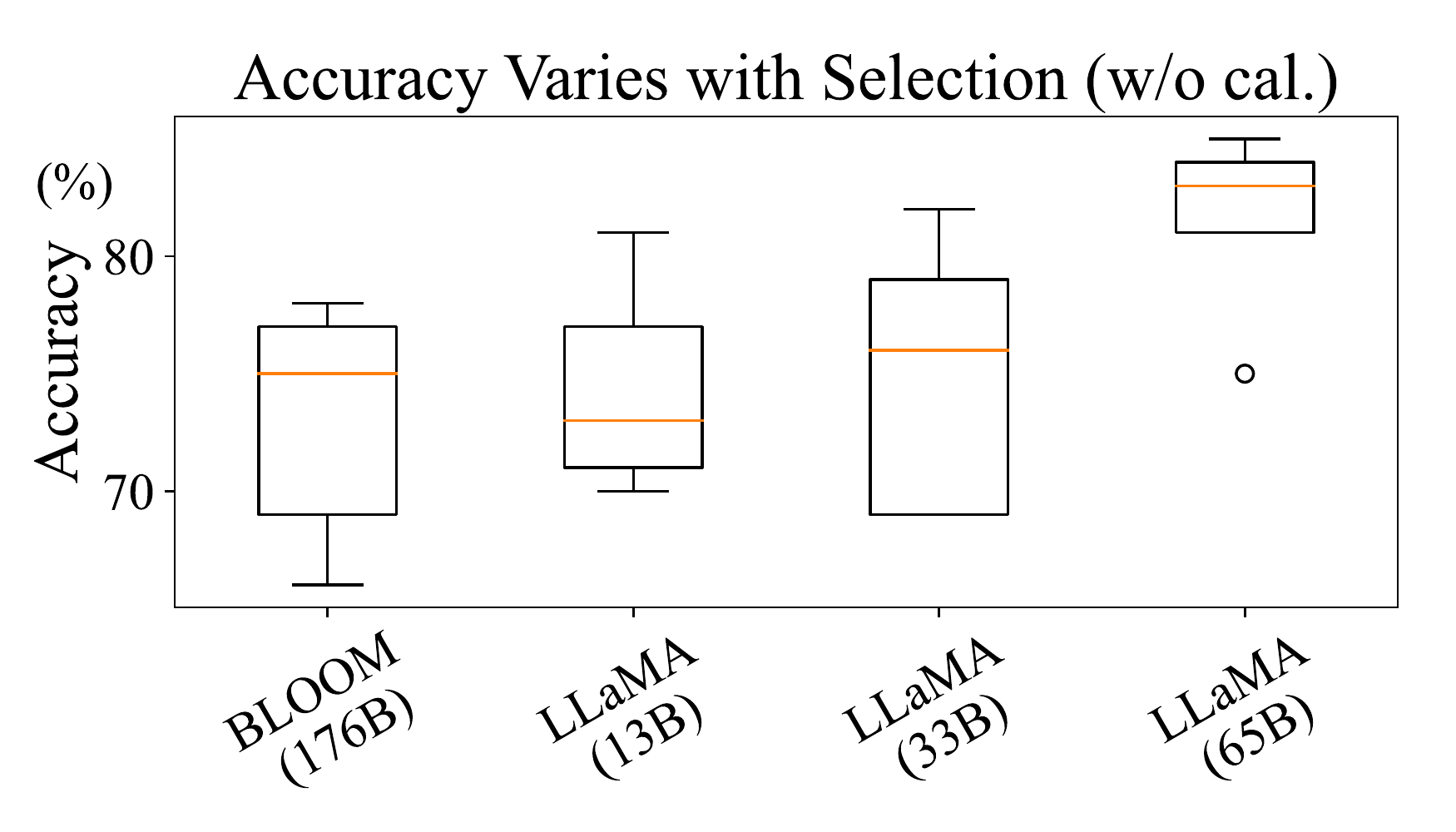}
    }  
    \subfloat[Selection (cal)]{
    \centering
    \includegraphics[width=0.24\linewidth]{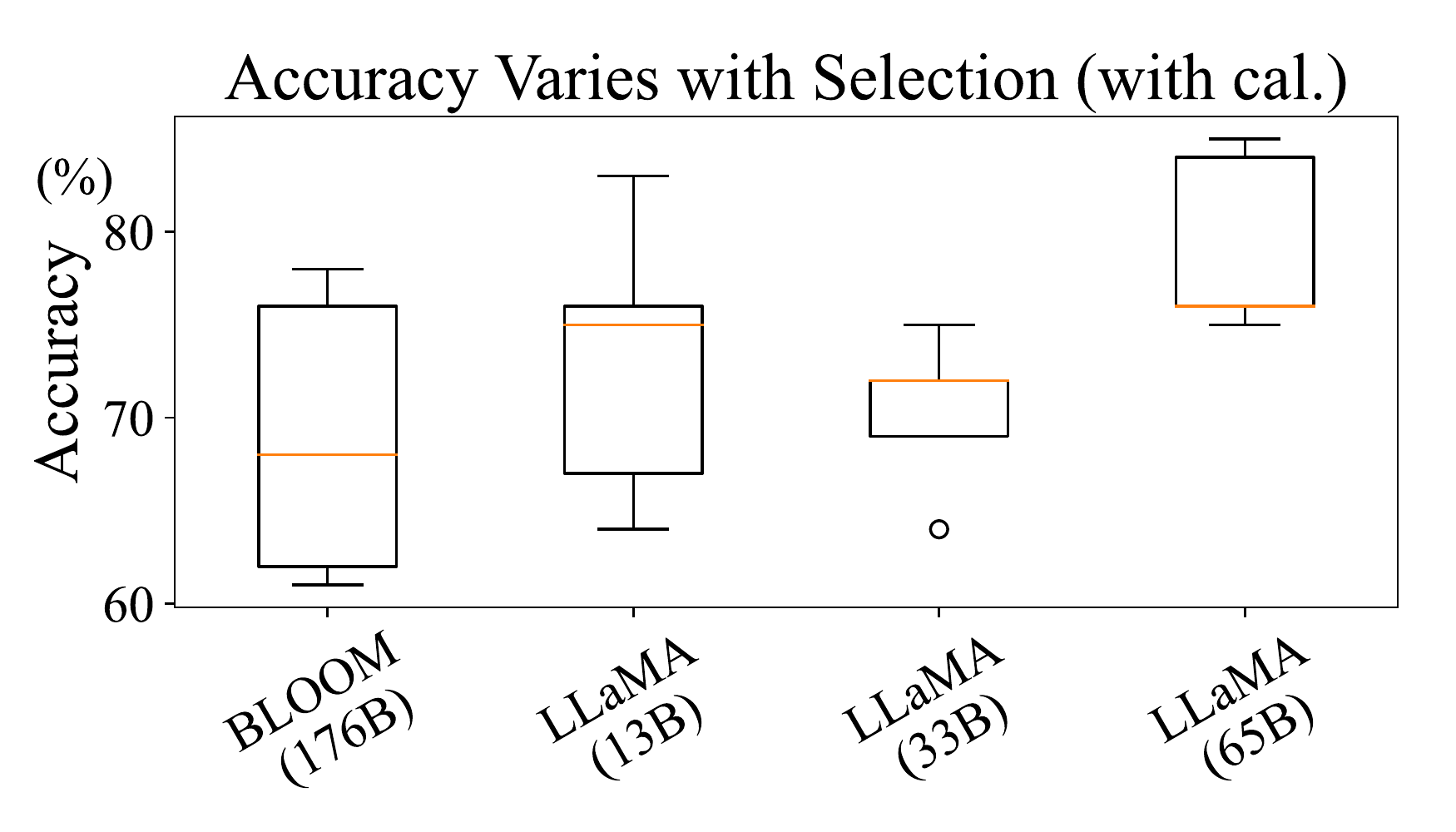}
    } 
    \subfloat[Permutation]{
    \centering
    \includegraphics[width=0.24\linewidth]{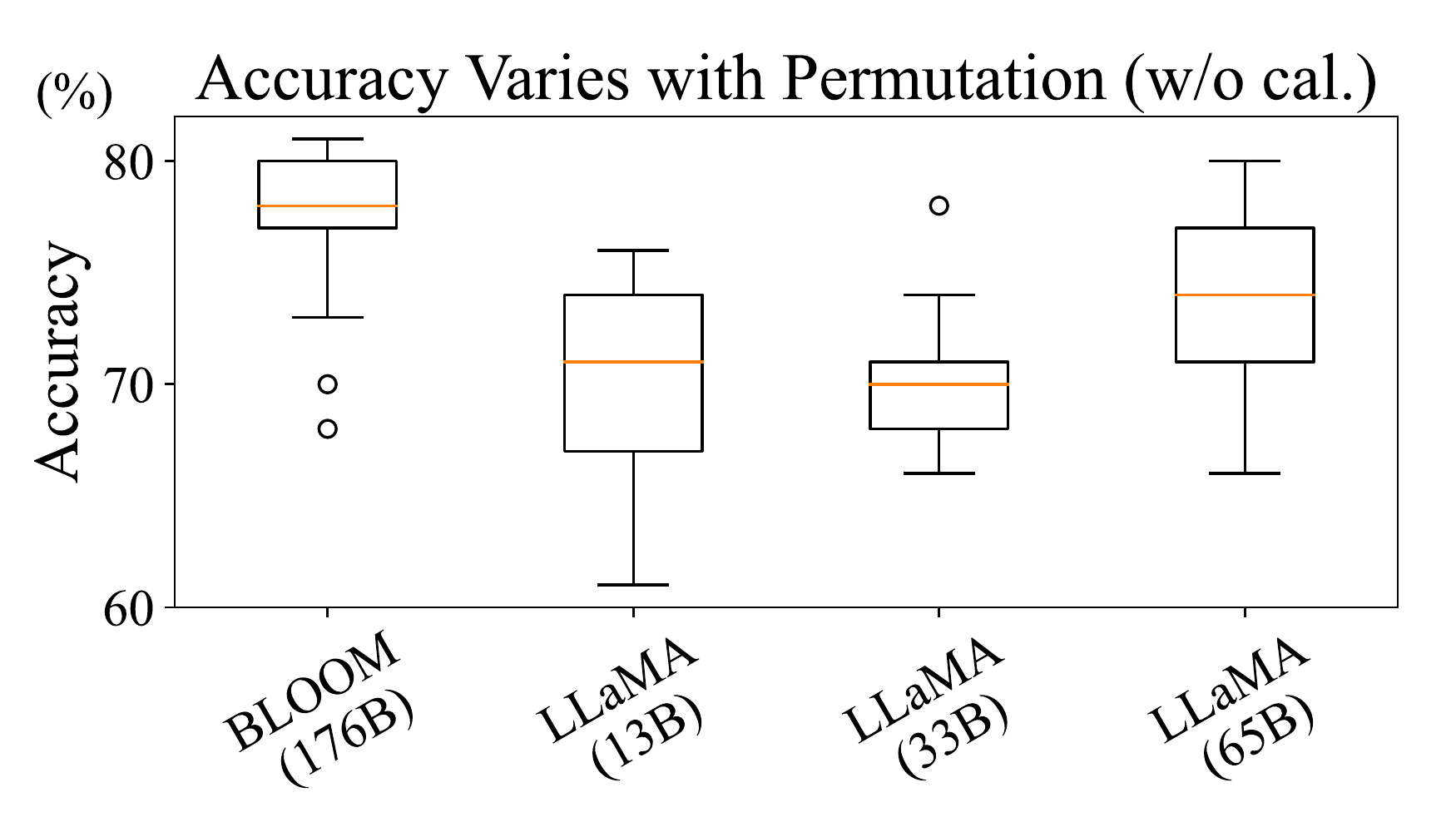}
    } 
    \subfloat[Permutation (cal)]{
    \centering    \includegraphics[width=0.24\linewidth]{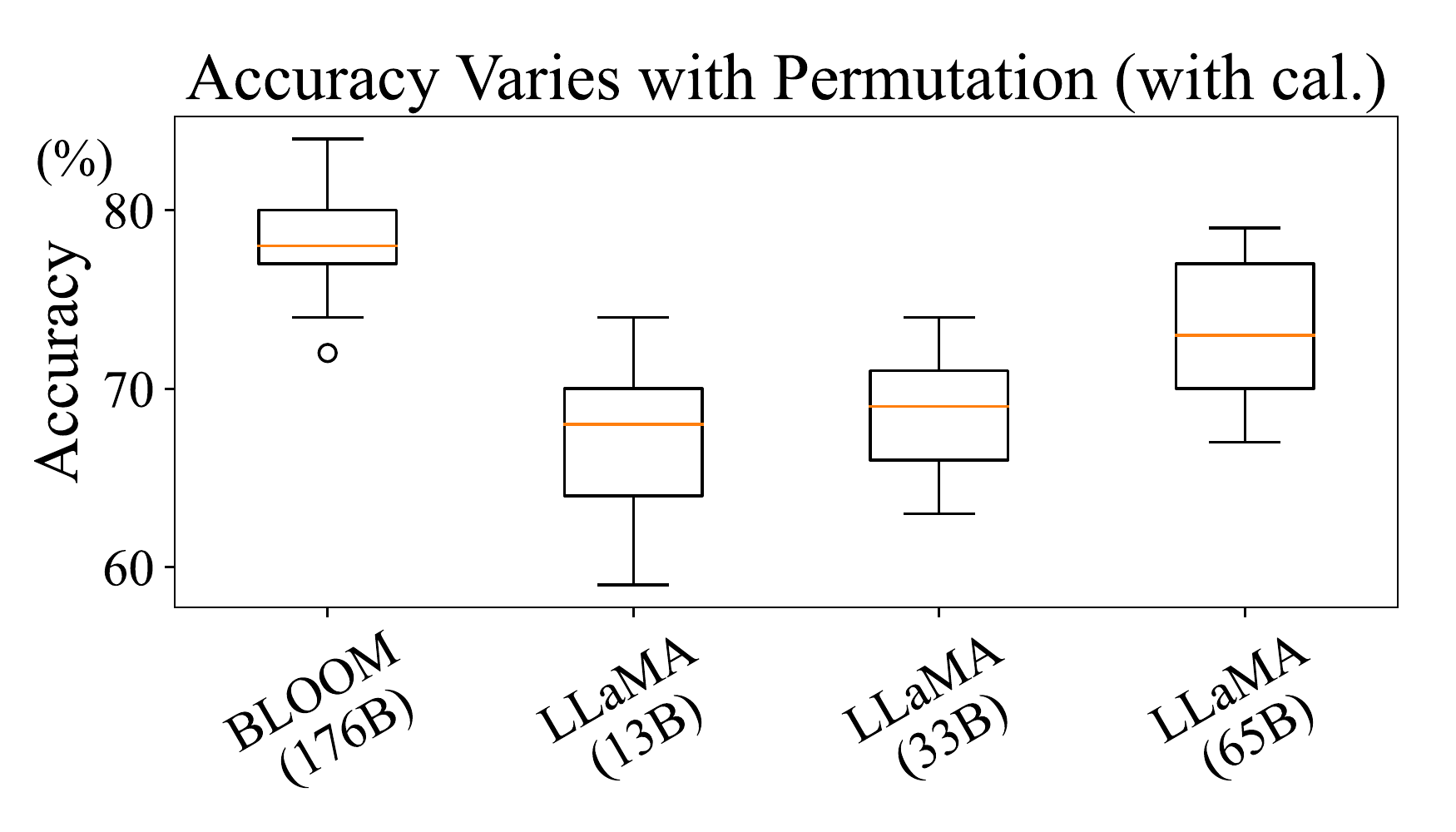}}
     
    \caption{ICL suffers from high instability due to high variations in demonstrations selection and order, even when  post calibration is performed.}
    \label{fig:obser-var}
\end{figure}


\textbf{Original space searching}\quad
A more intuitive approach for determining the best prompt is to search in the original space by selecting or reordering the prompt sentences entered by users. The searching can be concluded in two perspective. $\bullet$~\textbf{Global view}: A naive strategy is to \view{enumerate} all candidates to find the prompt that can achieve the best performance on validation set, but this strategy is computationally expensive since its complexity is $\sum_{k=1}^nC_n^kk!$. Zhang et al.~\cite{zhang2022automatic} find that errors frequently fall into the same cluster, where each cluster contains similar questions, so they proposed a \view{diversity-guided} searching strategy to select diverse demonstrations. In addition to demonstrations selection, \cite{order2021lu} have identified the impact of the prompt \view{order} on the results. They found the best sequence which yields the most diverse prediction results on the probing set by generating a probing set through LLMs. However, this method is also computationally expensive, and it may be difficult to ensure that the generated probing set is sufficiently balanced. $\bullet$~\textbf{Local view}: Previous studies~\cite{gentile2022fast} show that reducing the model’s \view{uncertainty} helps improve the model’s performance, and \cite{diao2023active} propose Active Prompting to select demonstrations according to the uncertainty of LLMs. KATE~\cite{liu2021makes} selects the prompt based on the \view{distance} amongst embeddings, with the goal of selecting the closest example. However, this method ignores the influence of the order of the examples and requires access to sentence embeddings.
\cite{shi2023large} demonstrate that LLMs can be easily distracted by irrelevant context, accordingly they identify several approaches for \view{filtering} out irrelevant information in context.

In the realm of original space searching, most of the current methods tend to focus solely on the influence of a singular factor (highlighted above) on performance, utilizing heuristic metrics to select context demonstrations that perform well according to this criterion. While these investigations certainly bring benefits to the community, they lack a comprehensive consideration of both local and global perspectives. The method proposed in this paper offers a metric to select context demonstrations from the perspective of predictive bias, which naturally facilitates a transition from the local view to global view.



%% file: secs/method.tex
\section{Revisiting the Sensitivity across Demonstrations}


In this section, we will clarify the notations and the templates used in this paper. Then, we will demonstrate some brief empirical results to show how different demonstration construction factors (e.g., example selection and order) affect performance. We further introduce the definition of predictive bias/fairness of a given prompt and show its connection to the predictive performance on different downstream tasks.

\subsection{Notations}
We consider a training set consisting of $N$ samples $S=\{(x_i, y_i)\}_i^N$, where $x_i$ is the sentence and $y_i \in \mathcal{Y}$ is the label of the $i^{th}$ training sample, and $\mathcal{Y}$ is the space of all labels for the task. We use a template $\Gamma(\cdot)$ to transform these sentences and labels into natural language space (i.e., prompt construction). Take an instance from the AGNews dataset~\cite{wang2018glue} for example, we have $x_i=\textit{"Cubans Risking Life for Lure of America."},~y_i=\textit{"World"}$, and $\Gamma(x_i,y_i)$ is $\textit{"Article: Cubans Risking Life for Lure of America. Answer: World"}$. We concatenate these demonstrations to form a prompt $\rho$, which by default is $\rho=\Gamma(x_1,y_1)\oplus\cdots\oplus\Gamma(x_n,y_n)$. At test time, we append the prompt $\rho$ with $\tau=\textit{"Article: <test sentence>. Answer: "}$ and feed it to a large language model $\mathcal{M}$. The predicted class is given by: 
\begin{equation} \hat{y}=\arg\max_{y \in \mathcal{Y}}\hat{p}(y|\rho \oplus \tau),\quad \hat{p}(y|\rho \oplus \tau)=\frac{\mathcal{M}(y|\rho \oplus \tau)}{\sum_{y \in \mathcal{Y}}\mathcal{M}(y|\rho \oplus \tau)},\end{equation} 
where $\mathcal{M}(y|\rho \oplus \tau)$ indicates the probability predicted by LLM, and the probability is normalized to fit the task. We denote the predictive distribution by $\hat{P}(x):=\{\hat{p}(y|\rho \oplus \tau)|y \in \mathcal{Y}\}$. In this paper, we focus on evaluating the instability caused by demonstrations, and we fix the prompt template following prior work~\cite{calibrate2021zhao}.

\begin{figure}[t]
    \centering
    \subfloat[AGNews (BLOOM 176B)]{
    \centering
    \includegraphics[width=0.31\linewidth]{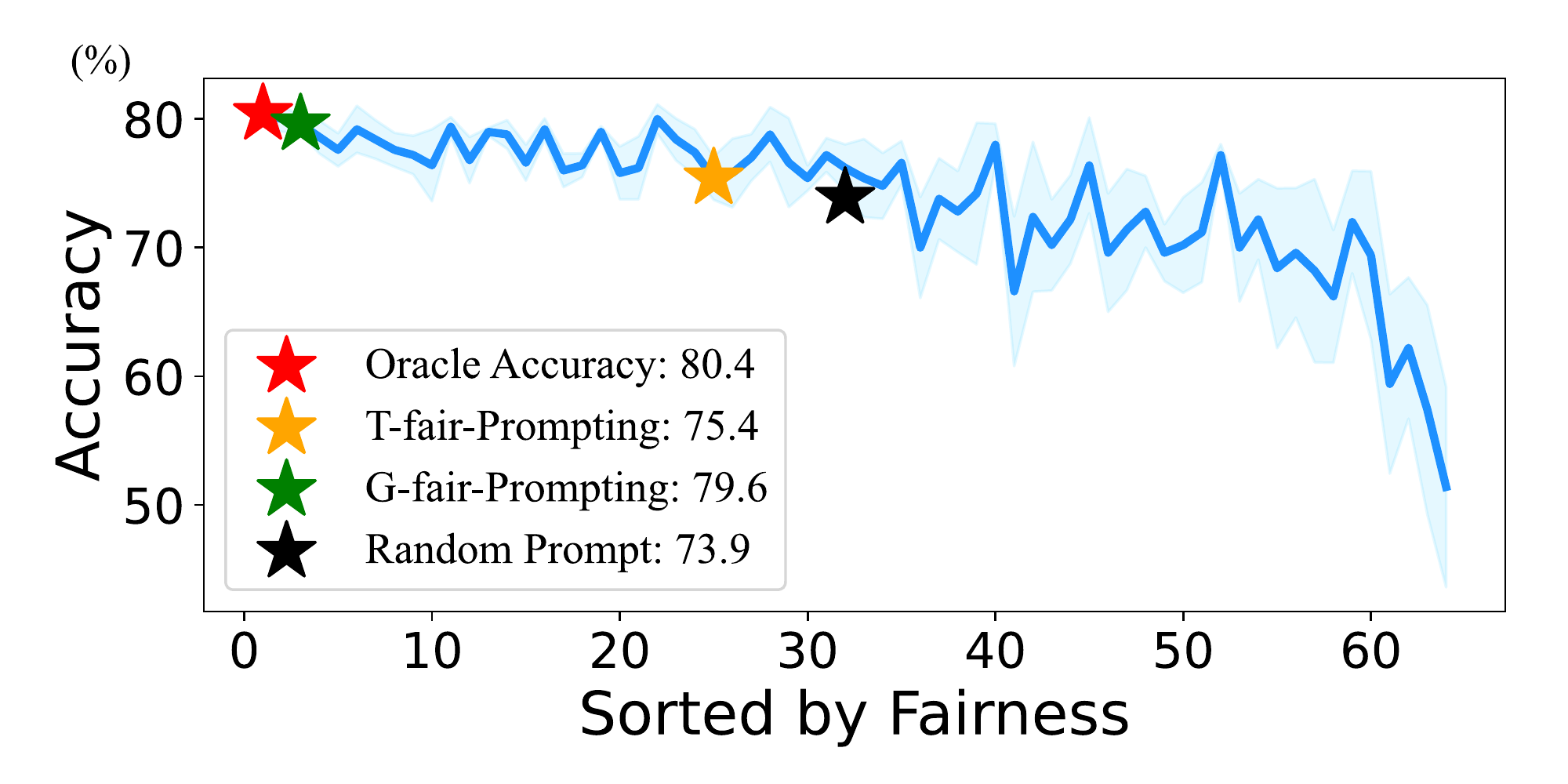}
    }  
    \subfloat[AGNews (LLaMA 13B)]{
    \centering
    \includegraphics[width=0.31\linewidth]{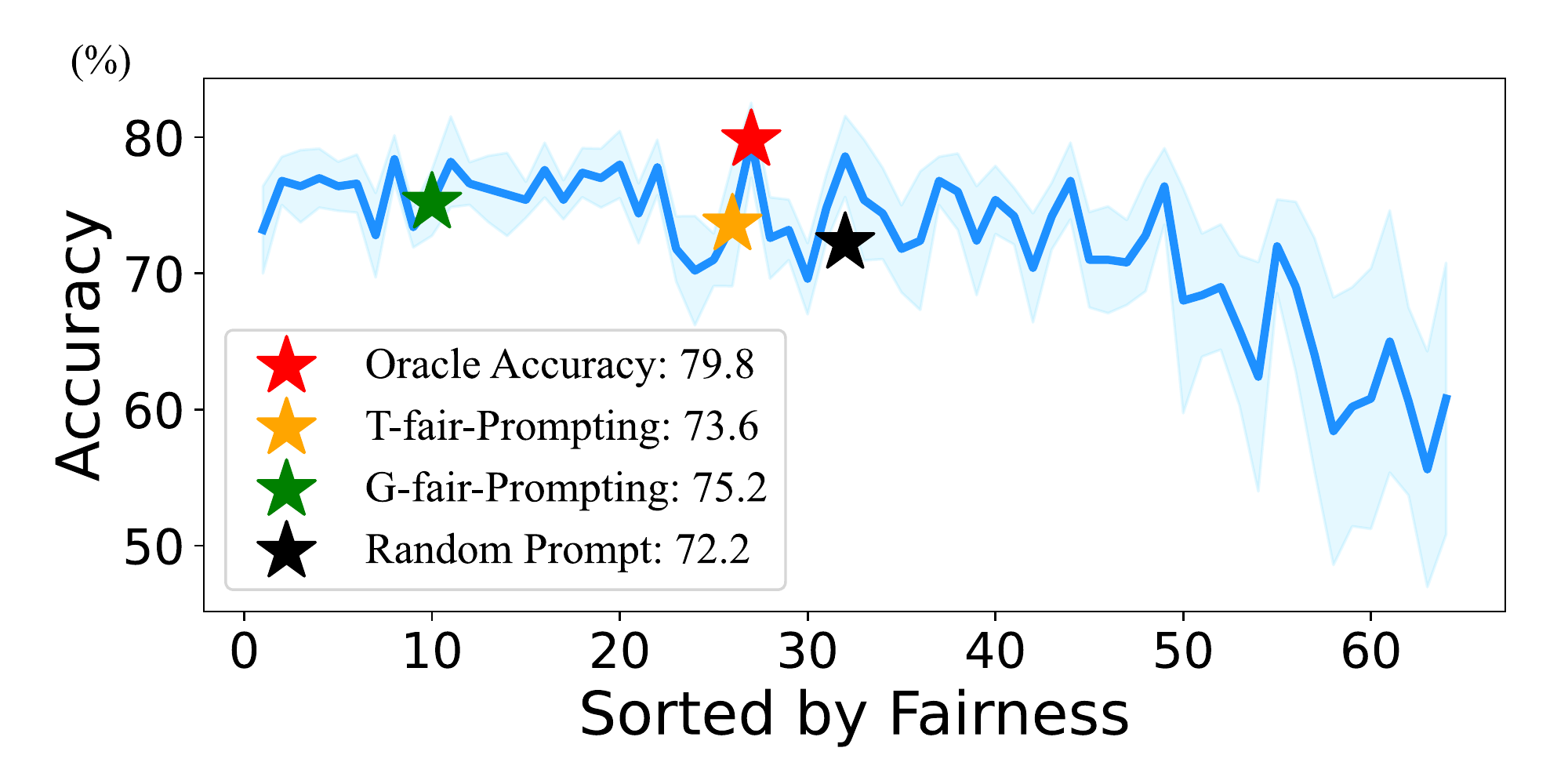}
    } 
    \subfloat[AGNews (LLaMA 65B)]{
    \centering
    \includegraphics[width=0.31\linewidth]{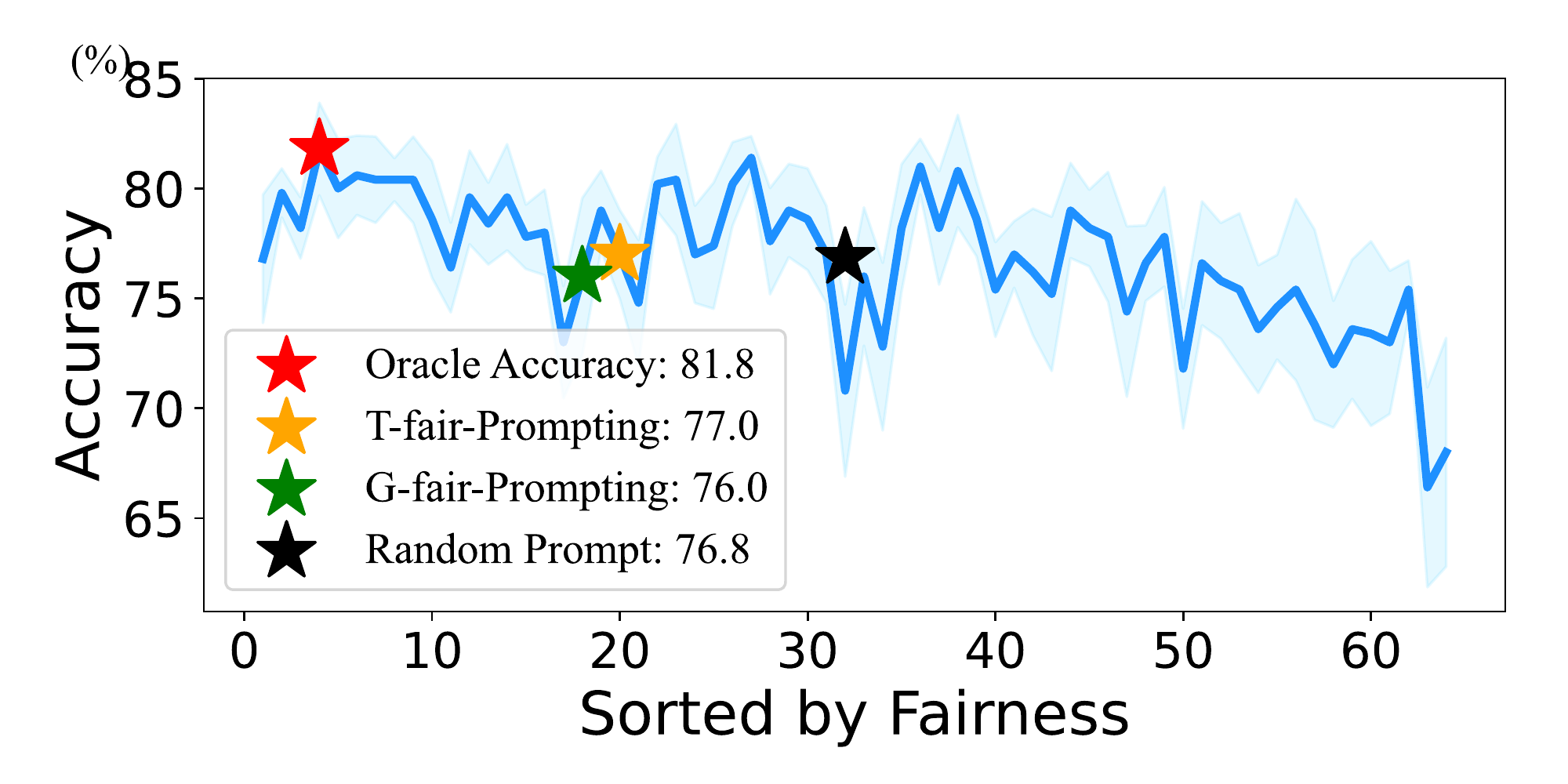}
    } \\ 
    \subfloat[TREC (BLOOM 176B)]{
    \centering
    \includegraphics[width=0.31\linewidth]{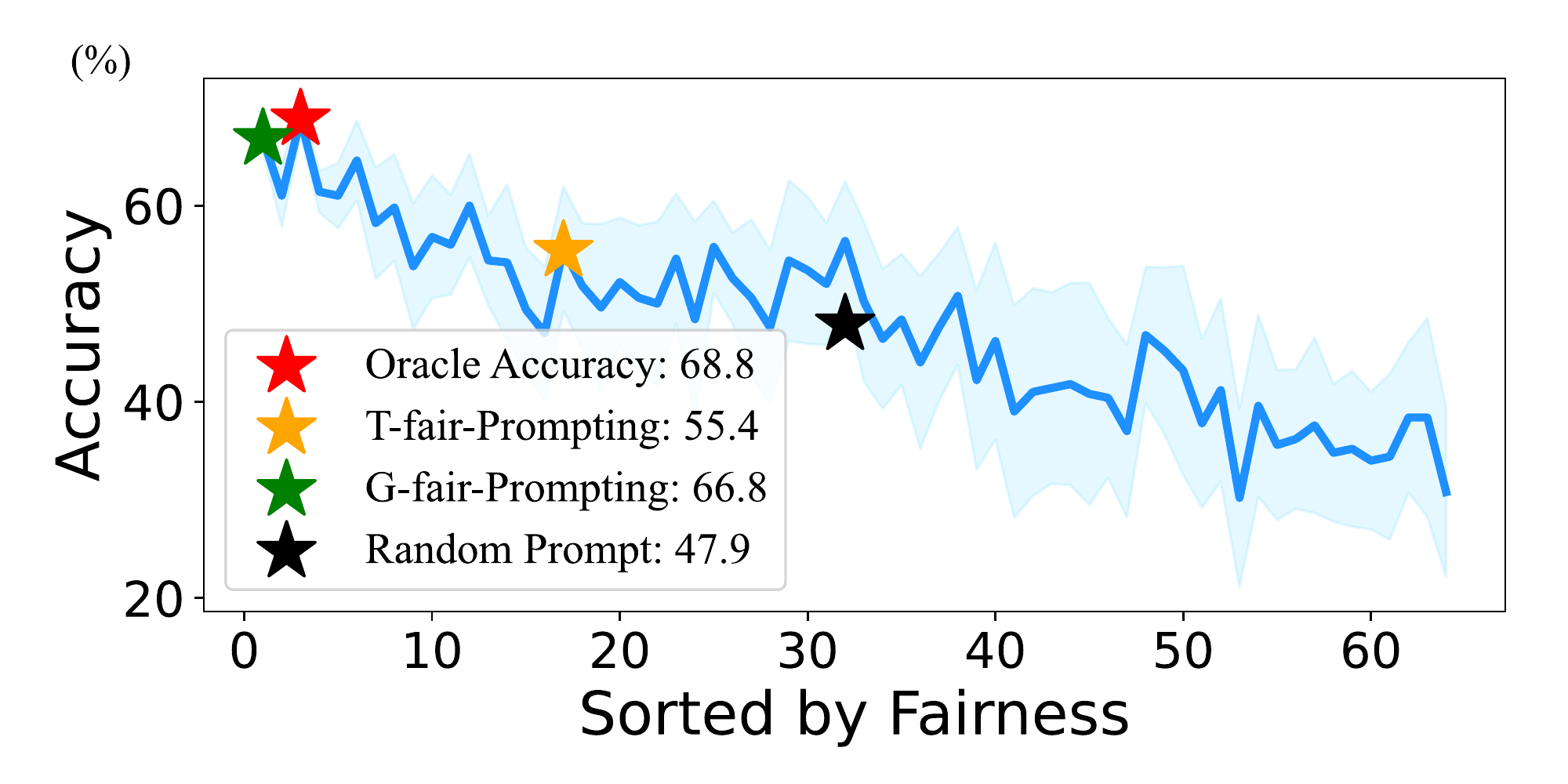}
    } 
    \subfloat[TREC (LLaMA 13B)]{
    \centering
    \includegraphics[width=0.31\linewidth]{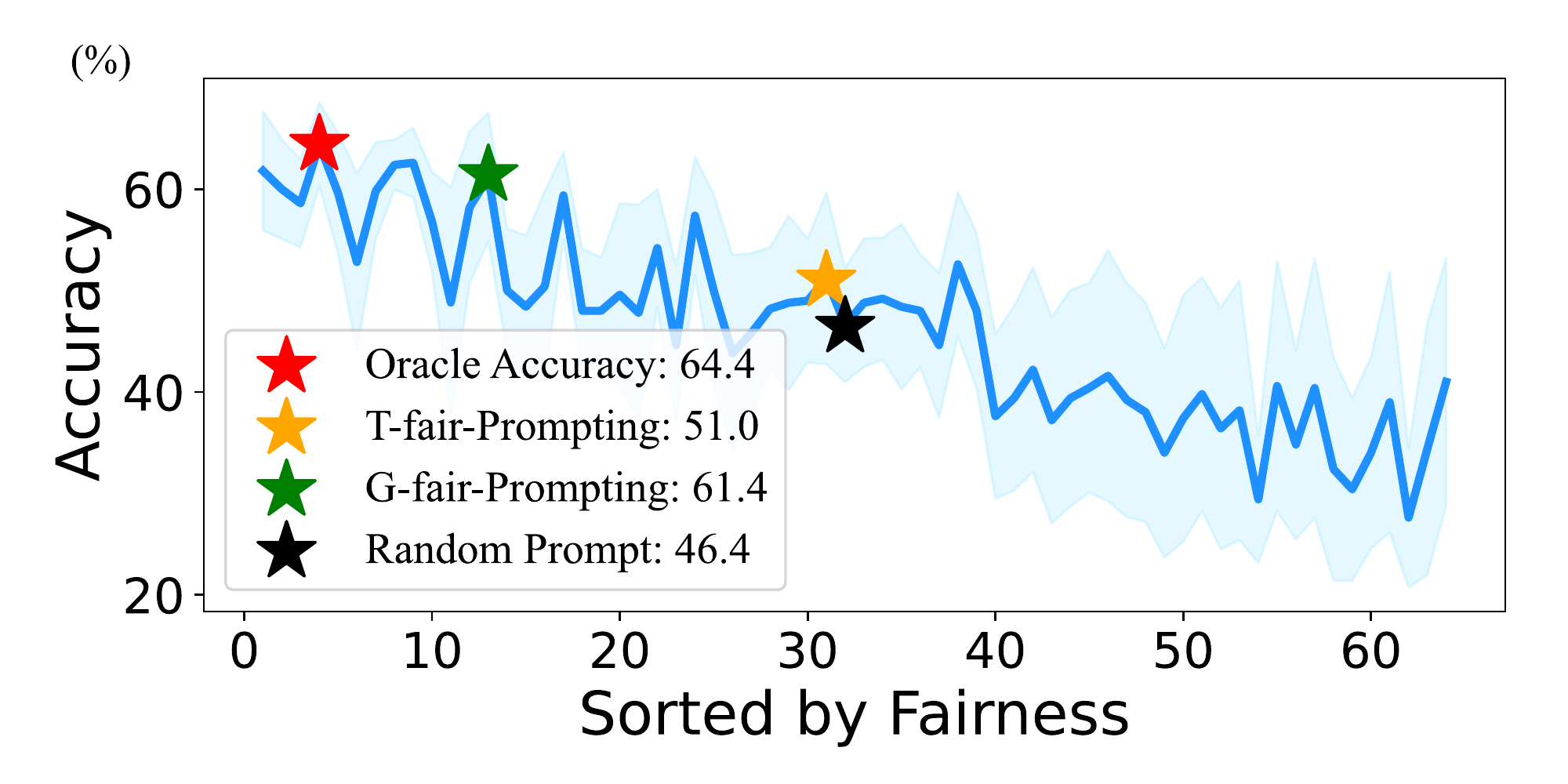}
    }
    \subfloat[TREC (LLaMA 65B)]{
    \centering
    \includegraphics[width=0.31\linewidth]{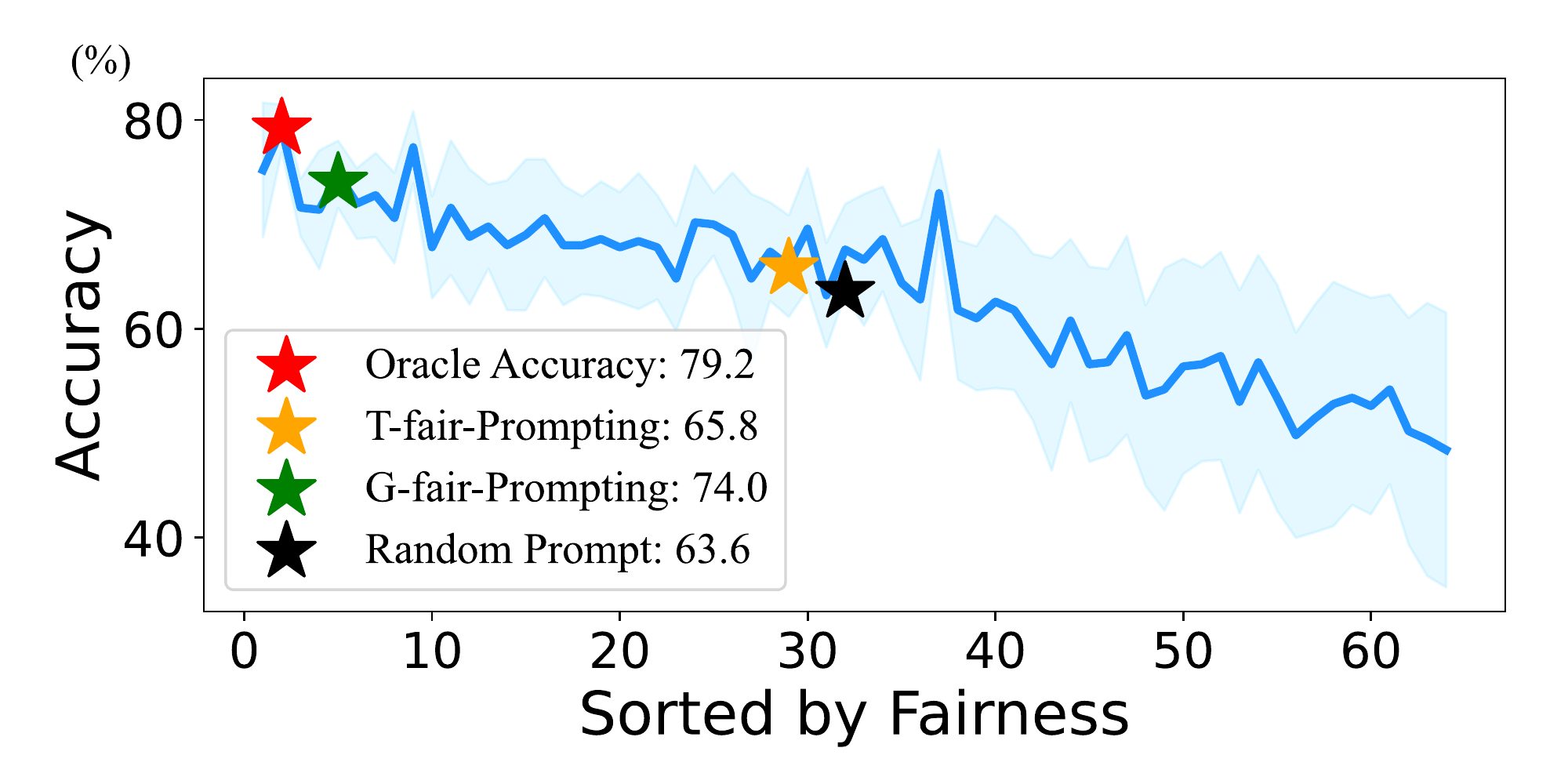}
    }
    \caption{Accuracy is highly consistency with fairness and greedy search can find a good prompt, where "Random" and "Oracle" indicate the average accuracy of all prompts and the upper-bound performance according to fairness.}
\label{fig:allcandidates}
\end{figure}

\subsection{Stability of Few-shot Prompting}
As demonstrated by prior research, the few-shot prompting technique is highly susceptible to a variety of factors, including the selection and order of demonstrations \cite{order2021lu,calibrate2021zhao}. In this study, we delve deeper into the stability of few-shot prompting, specifically focusing on the recently released LLaMA family by Meta \cite{touvron2023llama}. Additionally, we evaluate the stability of LLaMA models calibrated using the current state-of-the-art method \cite{zhang2022automatic,liu2021makes}.

To elucidate the impact of demonstration selection, we select four demonstrations for each different seed and randomly sample an order for each combination. Subsequently, we present the performance on AGNews in the form of a boxplot, which displays the data distribution based on a five-number summary (minimum, first quartile [Q1], median, third quartile [Q3], and maximum). As depicted in Fig.\ref{fig:obser-var}(a)(b), the accuracy demonstrates significant variability across various demonstrations.

To investigate the influence of permutations, we examine all possible permutations of four fixed demonstrations, resulting in $4!$ distinct candidates. Fig.\ref{fig:obser-var}(c)(d) also reveals a high degree of variance. While post-calibration contributes to mitigating instability, it is essential to note that the model remains sensitive even after post-calibration. This finding underscores the importance of meticulous demonstration selection. In subsequent experiments, we discover that our approach can be employed to further enhance the performance of the calibrated model.

\subsection{Predictive Bias of ICL}\label{sec:pre-bias}
As demonstrated in the preceding discussion, the performance of ICL is significantly impacted by various factors such as demonstration, permutation, and selection (refer to Appendix \ref{sec:app-ob} for additional information). Consequently, devising an efficient method for constructing an appropriate prompt with near-optimal performance is a crucial step in deploying LLMs for diverse downstream tasks. As outlined in the introduction, numerous studies aim to optimize prompts in ICL. This paper further investigates this issue through the lens of predictive bias, which refers to the discrepancy between targeted classes. \footnote{This notion differs slightly from the concept of social bias, which concentrates on specific feature attributes rather than labels. Our approach can be naturally extended to mitigate social bias in various settings.}

To achieve this, we initially introduce an efficient technique to assess the inherent predictive bias of a given prompt, drawing inspiration from previous work \cite{calibrate2021zhao}. We construct a training set-independent metric to measure predictive bias as follows: first, we merge the provided prompt with "semantic-free" test sample information (e.g., "[N/A]", denoted by $\eta$) and obtain the LLM's predictive distribution for this sample. Ideally, the predictive distribution should closely resemble a uniform distribution, as the test sample lacks semantic information. In this paper, we employ entropy as a measure of predictive bias, defined as:
\begin{equation}
    \fair(\rho) = -\sum_{y \in \mathcal{Y}} p(y|\rho \oplus \eta) \log p(y|\rho \oplus \eta)
    \label{eq:fair}
\end{equation}

Previous studies have utilized this metric to calibrate the model's output. In this paper, we conduct a comprehensive examination of the relationship between predictive bias and overall performance. Specifically, in a scenario with four training samples (due to the time-consuming nature of enumerating all prompt cases for a larger number), we enumerate all possible combinations and permutations of demonstrations for various datasets and LLMs. Subsequently, we arrange all candidates in descending order based on fairness, where an "index 0" denotes the prompt with the highest fairness. We perform experiments using five different seeds, resulting in training sets comprising distinct demonstrations while maintaining the test samples with seed 0. Fig.~\ref{fig:allcandidates} displays the results for different models, revealing a strong correlation between the model's performance and fairness score (i.e., fairer prompts yield better performance). The red star, referred to as the "Oracle" represents the optimal average performance, which consistently correlates with higher fairness. This observation prompts us to enhance the ICL performance by identifying the fairest prompt.

Nevertheless, discovering the fairest demonstration combination proves to be a formidable challenge, given the existence of $\sum_{k=1}^NC_N^kk!$ distinct candidates. As the size of the training set increases, this task becomes intractable. In order to tackle this problem, we propose two efficient strategies for approximating the most suitable demonstrations in the subsequent section.




\section{Fairest Prompt Search}

Drawing upon the aforementioned observations, we propose two strategies aimed at identifying the most fair prompt, which have been empirically demonstrated to achieve superior performance. Let us consider a training set $S$ comprising $n$ samples; the goal of these search strategies is to select a subset of samples from the training set and construct the context in a specific order so as to optimize the fairness criterion in Eq.~\ref{eq:fair}.

In an ideal scenario, we would consider the factors of demonstration selection and order permutation by examining $\sum_{k=1}^NC_N^kk!$ distinct candidates, which enumerates all possible situations. Here, $k$ represents the number of demonstrations selected, and $C$ signifies the combinatorial function. However, evaluating every candidate is infeasible, as demonstrated when $N=8$, yielding over $10^6$ candidates. In this paper, we introduce two search strategies to reduce computational cost: \topk and \greedy. The \topk strategy decreases complexity from $\Theta(\sum_{k=1}^NC_N^kk!)$ to $\Theta(N)$, but its performance hinges on the selection of $k$ and may be unstable when an unsuitable value of $k$ is chosen. As a result, we propose an additional greedy search strategy, termed \greedy, which lowers complexity to $O(N^2)$ and offers a superior approximation of the oracle solution.
Fig.~\ref{fig:app-cost} visualizes the computational costs over different training set size.

\begin{figure}[t]
    \centering
    \includegraphics[width=0.90\linewidth]{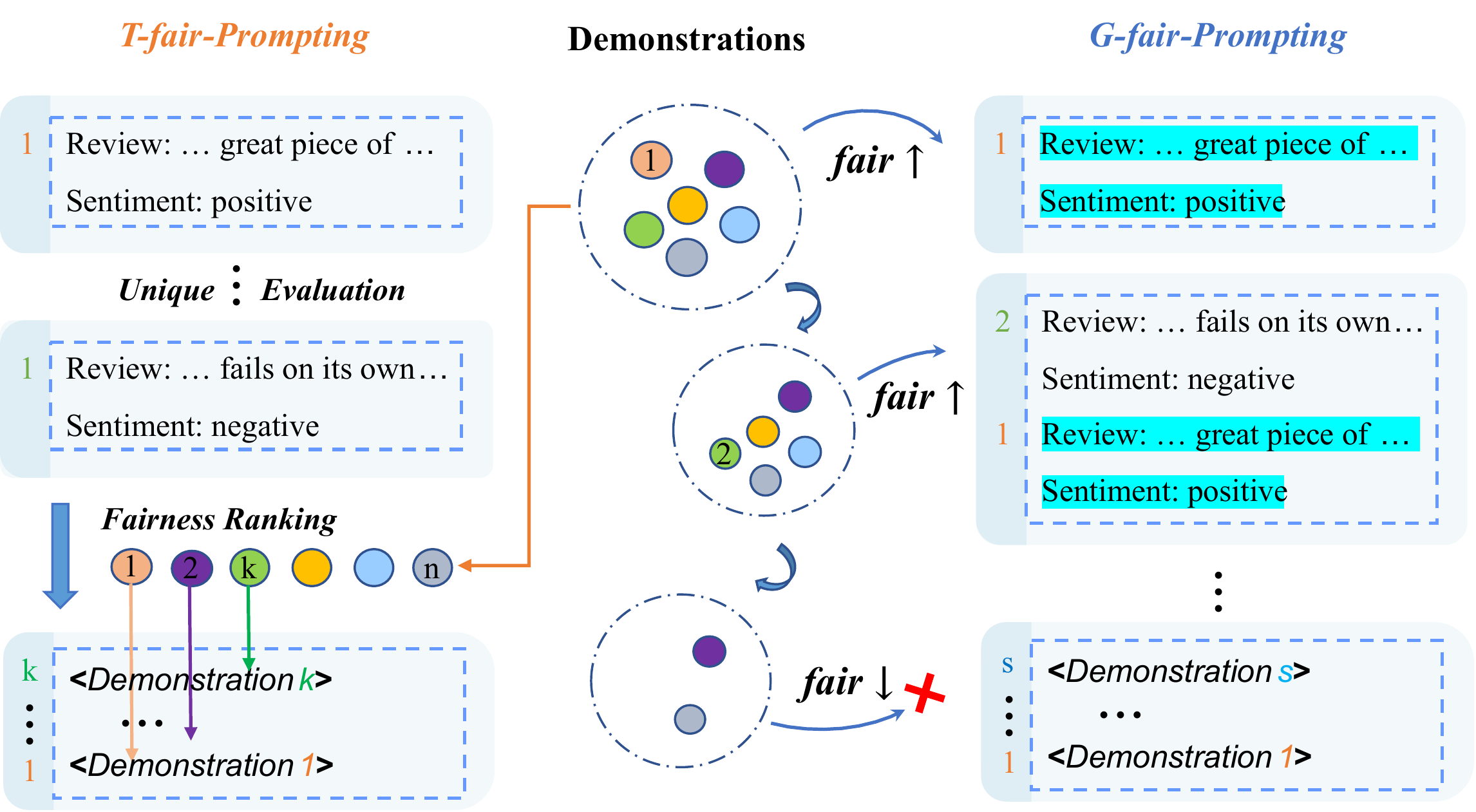}
    \caption{Overview of Most-fair Prompting.}
    \label{fig:method}
\end{figure}

\subsection{T-fair-Prompting}
The central idea of  \topk is founded on the heuristic understanding that the fairest prompt usually consists of demonstration samples with reduced individual biases. Consequently, \topk  constructs the prompt through a two-stage process. Initially, the prediction bias is assessed when the prompt is formulated using individual demonstrations. Subsequently, the top-$k$ fairest demonstrations are chosen and employed to prompt the LLM. It is important to note that fairer demonstrations are likely to be situated towards the end of the sequence, as the generation is more influenced by proximate demonstrations, in accordance with prior research~\cite{calibrate2021zhao}. A comprehensive description of the process is presented in Algorithm~\ref{alg:topk}, while a visual representation can be found in Fig.~\ref{fig:method}. Specifically, when $k$ is equivalent to the size of the training set, the method degrade to a search for the optimal order of demonstrations. Nevertheless, \topk  is heavily reliant on the chosen value of $k$. More crucially, \topk addresses this issue through a purely local perspective, thereby neglecting considerations from a global standpoint, which typically results in sub-optimal outcomes. As a result, we subsequently introduce the \greedy method, which operates in a local-to-global fashion, as described below.

\subsection{G-fair-Prompting}
The \greedy algorithm adheres to the standard procedure of greedy search, which seeks the optimal solution by making locally optimal choices at each stage. In each step of the algorithm, the chosen demonstration is the one that allows the updated prompts to achieve the highest fairness score. This strategy balances the quality of the search with the worst-case time complexity. By accepting an increased worst-case time complexity of $O(N^2)$, the search quality is significantly enhanced. It is important to note that the \greedy algorithm operates from a local to global perspective as shown by Algorithm. During the initial stages, the bias of individual samples is taken into account, while the later stages focus on reducing global predictive bias. Specifically, at each step, we insert a new demonstration $\Gamma(x_i,y_i)$ from the remaining demonstration set $\mathcal{S}'$ (ensuring demonstrations are not repeated) at the beginning of the current context $\rho$ and select the demonstration that maximizes the  fairness improvement. Formally, at step 9 in Algorithm~\ref{alg:greedy}, the inserted demonstration should satisfy the following criterion:
\begin{equation}
    \arg \max_{x_i\in \mathcal{S}'} \fair(\Gamma(x_i,y_i)\oplus\rho)\quad \text{s.t. }\fair(\Gamma(x_i,y_i)\oplus\rho)>\fair(\rho).
\end{equation}


\begin{minipage}{0.45\textwidth}
\begin{algorithm}[H]
    \caption{\topk}
    \label{alg:topk}
    \begin{algorithmic}[1] 
        \STATE \textbf{Given:} training set $S=\{(x_i, y_i)\}_i^N$, pretrained LLM $\mathcal{M}$, transformation template $\Gamma(\cdot)$, and context-free input $\eta$
        \STATE Initial prompt $\rho$
        \FOR{$(x_i, y_i)$ in $S$}
        \STATE Inference $\hat{P} \leftarrow \{\hat{p}(y|\Gamma(x_i,y_i)\oplus\eta)|y \in \mathcal{Y}\}$ via $\mathcal{M}$
        \STATE Calculate the $\fair(\Gamma(x_i,y_i))$ according to Eq.~\ref{eq:fair}       
        \ENDFOR
        \STATE Sort $\fair_{i=1,\cdots,N}(\Gamma(x_i,y_i))$ in descending order
        \FOR{$d$ in $1, \cdots, k$}
        \STATE \emph{Insert} the most $d$ fair demonstration at the head of $\rho$
        \ENDFOR
        \RETURN $\rho$

    \end{algorithmic}
\end{algorithm}
\end{minipage}
  \hfill
\begin{minipage}{0.53\textwidth}
\begin{algorithm}[H]
    \caption{\greedy}
    \label{alg:greedy}
    \begin{algorithmic}[1] 
        \STATE \textbf{Given:} training set $S=\{(x_i, y_i)\}_i^N$, pretrained LLM $\mathcal{M}$, transformation template $\Gamma(\cdot)$, and context-free input $\eta$
        \STATE Initial prompt $\rho$
        \WHILE{$S$ \emph{is not null}}
        \FOR{$(x_i, y_i)$ in $S$}
        \STATE $\rho_\text{tmp} \leftarrow  \Gamma(x_i,y_i)\oplus\rho$
        \STATE Inference $\hat{P} \leftarrow \{\hat{p}(y|\rho_\text{tmp}\oplus\eta)|y \in \mathcal{Y}\}$ via $\mathcal{M}$
        \STATE Calculate the $\fair(\rho_\text{tmp})$ according to Eq.~\ref{eq:fair}
        \ENDFOR
        \STATE \emph{Insert} the demonstration that can improve fairness best and \emph{remove} it from $S$ 
         \STATE \emph{Stop} searching when fairness can't be improved
        \ENDWHILE
        \RETURN $\rho$
 
    \end{algorithmic}
\end{algorithm}
\end{minipage}



%% file: secs/exp.tex
\section{Experiments}
\subsection{Experimental Setup}
\textbf{Models.} There are a large number of available LLMs (Appendix~\ref{sec:app-models}) including open-source models and black-box cloud API. Recently, Meta has released their powerful pretrained LLMs, LLaMA. LLaMA models with 13B parameters can achieve comparable performance in contrast to BLOOM and GPT-3 with much larger model size. In this paper, we evaluate the effectiveness of our method on BLOOM (176B) and LLaMA models of different sizes. We have opted to employ LLaMA (65B) as a substitute for GPT-3 in our experiments, since oepnai strictly restricts the API access to certain areas.

\textbf{Datasets.} We conducted experiments on various text classification datasets~\cite{wang2018glue}, namely SST-2, AGNews, CoLA, TREC, and RTE. Furthermore, the maximum input length of LLaMA is 512, and the sentences in RTE are too long for LLaMA. The task descriptions and statistics are available in Table~\ref{tab:datasets}.
\input{tabs/datasets}

\subsection{Results}
\input{tabs/accuracy}
We conducted experiments on different settings and reported the results of five runs. We compared our method with the diversity-guided searching strategy proposed by Zhang et al.\cite{zhang2022automatic} (Global view) and the similarity-guided searching strategy proposed by Liu et al.\cite{liu2021makes} (Local view). Note that methods based on local view are time-consuming since they require searching different demonstrations for every test example. Table~\ref{tab:four-topk-greedy} shows the performance of the different strategies, where "Random" indicates the average accuracy for enumerating all situations, "Diversity" and "Similarity" indicate demonstrations are selected according to diversity and similarity, respectively. For each dataset, we set the size of the training set to 4. "Diversity" and "Similarity" select 4 from 16 demonstrations, as they need more candidates. The baseline is expensive to compute since enumerating all candidates for 4 demonstrations in RTE on BLOOM will take more than 120 NVIDIA A100 GPU hours. We enumerate all candidates for the training set with 4 demonstrations on different models, as shown in Fig.~\ref{fig:allcandidates}. The results on models whose parameters less than 13B are shown in Table~\ref{tab:app-acc} (i.e., GPT2-XL (1.5B), LLaMA (7B), and LLaMA (13B)).

$\bullet$ \textbf{\greedy can reach a close approximation of enumeration.} To evaluate whether the \greedy (Greedy) method can approximate the best performance of enumerating all candidates, we marked the performance of \greedy with a green star (representing the closest value to averaged accuracy of \greedy on the line). We found that \greedy can achieve a very close approximation to enumeration. As shown in Fig.~\ref{fig:allcandidates}, most prompts searched by \greedy achieved a top $20\%$ ranking, and on BLOOM (176B), \greedy almost found the most fair prompt.

$\bullet$ \textbf{\greedy outperforms \topk.} As shown in Table~\ref{tab:four-topk-greedy}, although \topk achieves better performance compared with random selection, \greedy consistently outperforms \topk. Furthermore, Top-2 significantly outperforms Top-4 in most cases (over $5\%$), indicating that the number of demonstrations selected is crucial. Overall, the results demonstrate that \greedy achieves satisfactory performance with only a slight additional cost.

\begin{wrapfigure}{r}{0.45\textwidth}
\vspace{-0.6cm}
\centering
  \includegraphics[width=0.90\linewidth]{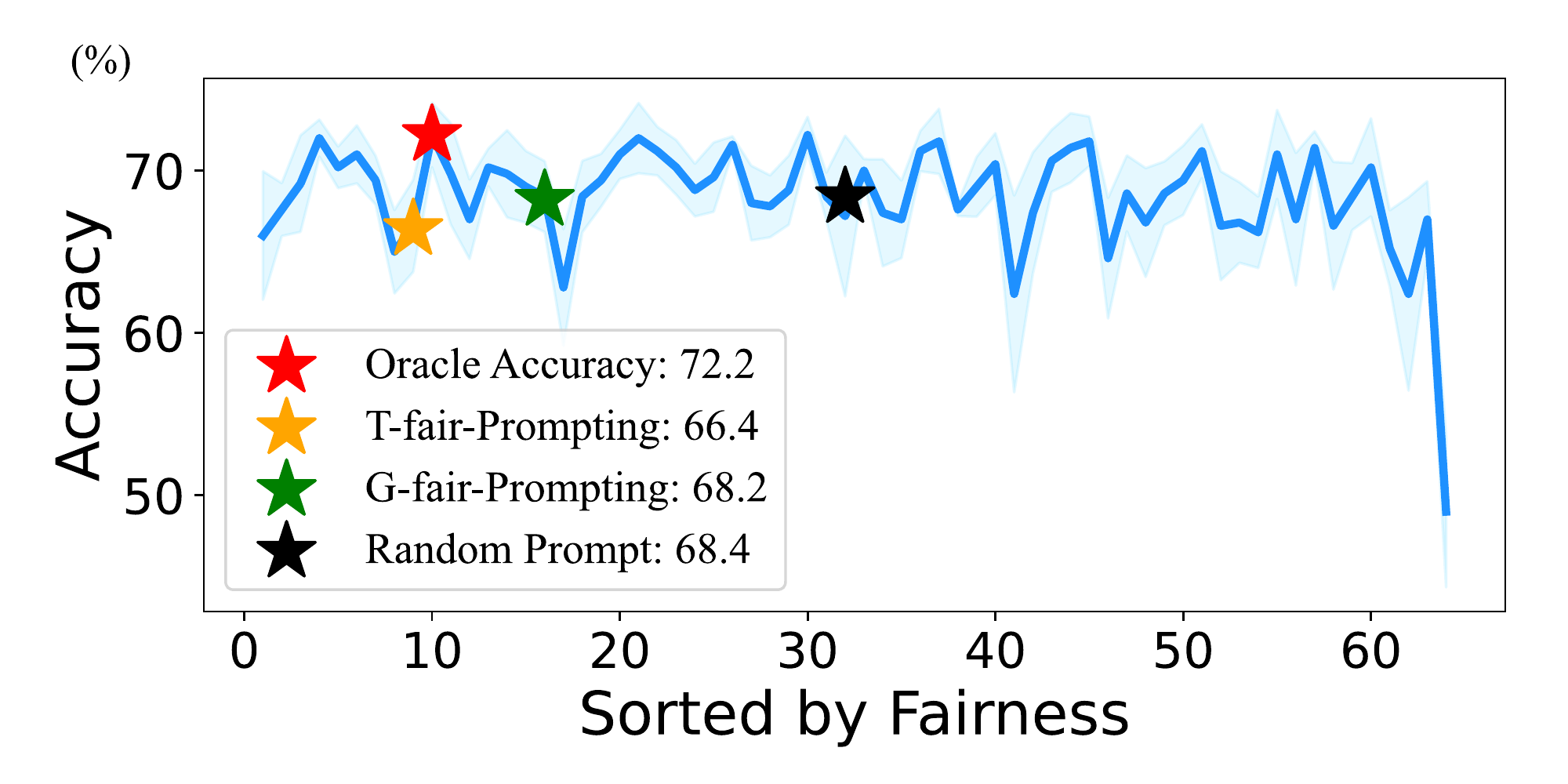}
 \caption{BLOOM is not sensitive to CoLA.}
 \vspace{-0.3cm}
 \label{fig:cola-bloom}
\end{wrapfigure}
$\bullet$ \textbf{Compared with SOTA methods.} We compared our methods with several State-of-the-Art (SOTA) methods, including diversity-guided and similarity-guided techniques. We observed that our \textbf{greedy} approach outperforms most of these SOTA methods in most situations, and the improvements of over $10\%$ are observed on dataset TREC. The similarity-guided method, on the other hand, achieved the best performance on the topic classification task (AGNews). This is because it searches for a unique prompt for every different test example based on the distance between the embeddings of the training samples and the test example. This strategy selects demonstrations with labels that are the same as the test samples, and Language Models (LLMs) tend to predict biased predictions toward the labels that always appear in the context. However, the similarity-guided method may prove inadequate when applied to other tasks. Specifically, the similarity-guided strategy exhibits lower performance compared to random selection in QC and acceptability tasks. Furthermore, the \greedy approach may occasionally falter when the model's sensitivity to the task is not immediately evident, as observed in the acceptability task on BLOOM (depicted in Fig.~\ref{fig:cola-bloom}). Note that the training set size of compared methods is $4\times$ larger than ours.

$\bullet$ \textbf{Comparison with Calibration Method.} Post-calibration~\cite{calibrate2021zhao}, can enhance the accuracy of a given prompt in most cases. However, when the selected prompt is of poor quality, the performance may remain inadequate even after calibration. We compared the performance of \greedy with random selection with calibration (averaged on all candidates), and found that \greedy can outperform random selection with calibrated most situations. For example, on the topic classification task, \greedy achieves the best performance on most models. Moreover, we find that post calibration can harm the performance of the model and it occurs significantly times, so it is worthwhile to reconsider the influence of manipulating the model's probability directly.

\input{tabs/calibrated}
Post calibration~\cite{calibrate2021zhao} can improve the accuracy of a certain prompt (in most cases), but when the selected prompt is very poor, the performance still very poor even after calibration. We conducted experiments (Table~\ref{tab:calibrated}) to compare the performance of \greedy and random selection with calibration ("Average" and "Worst" indicate averaged accuracy and worst performance on all permutations of training examples), and observed that \greedy outperforms random selection with calibration in most case. For instance, on the CoLA, \greedy exhibited superior performance on most models. Additionally, we find that post-calibration could negatively affect the model's performance in many scenarios while it sometimes can improve the performance significantly even on selected prompts, for example, an improvement by $10\%$ is observed on BLOOM-TREC. Hence, it is crucial to reconsider the impact of directly manipulating the model's probability.

%% file: tabs/datasets.tex
\begin{table}[ht]
\centering
\begin{threeparttable}[b]
\caption{Dataset descriptions.}
\label{tab:datasets}
\begin{tabular}{ccccc} \toprule
{\textbf{Corpus}}                  & {\textbf{Task}} &\textbf{Classes}& \textbf{Domain} & \textbf{Total Cost}\tnote{1}           \\ \midrule {SST-2}  &  sentiment & 2 &  movie reviews & over $60$ GPU hours  \\ {TREC}  &  QA/QC & 6 &  open domain & over $220$ GPU hours \\ {AGNews}  &  topic & 4 & news &over $250$ GPU hours \\ {CoLA}  &  acceptability & 2 &  misc. & over $160$ GPU hours  \\  {RTE}\tnote{2}  &  NLI & 2 &  news, Wikipedia &over $110$ GPU hours  \\ \bottomrule
\end{tabular}
\begin{tablenotes}
     \item[1] Total Cost$=$Hours$\times$GPUs. Hardware: BLOOM$=$A100, LLaMA$=$V100. 
     \item[2] Not applicable to LLaMA because of the maximum prompt token limit.
   \end{tablenotes}
   \end{threeparttable}
\end{table}

%% file: tabs/accuracy.tex
\begin{table}[ht]
\centering
\caption{Accuracy for different prompting strategies (averaged on $5_{0,\cdots,4}$ different seeds, where Top-$k$ and Greedy indicate \topk with $k$ demonstrations and \greedy respectively).}
\label{tab:four-topk-greedy}
\resizebox{1.0\textwidth}{!}{\begin{tabular}{c|c|ccc||ccc} \toprule
\multirow{2}{*}{\textbf{Model}}                  & \multirow{2}{*}{\textbf{Dataset}}                  & \multirow{2}{*}{\textbf{Random}} & 
 \multirow{2}{*}{\textbf{Diversity}} & \multirow{2}{*}{\textbf{Similarity\tnote{1} }} & \multicolumn{3}{c}{\textbf{Ours}} \\ & & & &  &\textbf{Top-2} & \textbf{Top-4} & \textbf{Greedy}                  \\ \midrule \multirow{7}{*}{BLOOM (176B)} & {SST2}   & $92.7_{2.3}$&$\best{95.0_{0.9}}$ &$94.0_{0.9}$   & ${94.6_{0.5}}$      & $93.8_{2.1}$     & ${91.2_{4.0}}$  \\ \cmidrule{2-8} &    {AGNews}                            & $73.9_{5.9}$  & $70.2_{10.1}$ &$74.8_{3.8}$  &$75.4_{2.2}$      & $74.8_{2.3}$       & $\best{79.6_{1.4}}$     \\ \cmidrule{2-8}
                                & {TREC}   & $47.9_{14.6}$&$46.0_{8.7}$ &$31.4_{3.1}$   &$55.4_{13.3}$            & $39.2_{19.3}$     & $\best{66.8_{2.5}}$  \\ \cmidrule{2-8}
                                & {RTE}  & $62.4_{4.2}$ &$\best{69.2_{1.9}}$&$67.2_{3.5}$  &$55.6_{1.0}$            & $57.6_{1.9}$      & ${63.0_{2.1}}$ \\  \cmidrule{2-8}
                                & {CoLA}   & {$68.4_{4.8}$} & \best{$71.0_{3.7}$} & $69.8_{2.5}$ &$66.4_{8.6}$            & $66.8_{3.7}$       & ${68.2_{6.2}}$      \\  \midrule\multirow{5}{*}{LLaMA (33B)}& {SST2}   & $82.5_{11.8}$&$\best{90.0_{2.7}}$ &$72.8_{4.4}$   & ${82.0_{11.1}}$      & $80.0_{12.2}$     & ${85.6_{8.2}}$  \\ \cmidrule{2-8} &   {AGNews}           & {$75.2_{5.0}$}            &$75.0_{5.1}$      & {$75.0_{2.4}$}        &$73.2_{3.9}$      & $69.8_{4.4}$            & $\best{76.4_{4.6}}$    \\ \cmidrule{2-8}
                                & {TREC} & $68.1_{11.1}$ &$68.2_{4.7}$            & $60.6_{3.4}$ &$71.4_{11.1}$            & $57.8_{17.3}$        & $\best{80.2_{5.3}}$  \\ \cmidrule{2-8}
                                & {CoLA}   & $66.9_{11.0}$ & $68.8_{6.8}$ & $72.8_{2.0}$ &$63.8_{13.3}$            & $69.8_{3.9}$       & $\best{70.6_{4.2}}$      \\ \midrule\multirow{5}{*}{LLaMA (65B)} & {SST2}   & $90.0_{7.7}$&$90.8_{9.0}$ &$87.4_{3.1}$   & ${88.2_{8.6}}$      & $\best{95.8_{1.5}}$     & ${87.8_{9.0}}$  \\ \cmidrule{2-8}
                                &    {AGNews}                          & $76.8_{5.0}$   &$\best{78.2_{3.1}}$      & \best{$78.2_{1.8}$}  &${77.0_{3.4}}$      & $76.2_{4.9}$           & $76.0_{4.0}$    \\ \cmidrule{2-8}
                                & {TREC} & $63.6_{14.2}$ &$65.2_{10.9}$            & $64.0_{5.5}$   &$65.8_{13.0}$            & $57.4_{19.9}$       & $\best{74.0_{12.2}}$  \\ \cmidrule{2-8}
                                & {CoLA} & $66.2_{9.8}$ &$62.6_{8.6}$& $59.2_{14.0}$  &$67.6_{11.7}$            & $62.6_{6.5}$       & $\best{72.0_{4.5}}$      \\ \bottomrule
\end{tabular}}
\end{table}

%% file: tabs/calibrated.tex
\begin{table}[ht]
\centering
\caption{Accuracy comparison after post calibration.}
\label{tab:calibrated}
\begin{tabular}{c||c|cccccc} \toprule
\multirow{2}{*}{\textbf{Dataset}}                  & \multirow{2}{*}{\textbf{Method}}    & \multicolumn{2}{c}{\textbf{BLOOM (176B)}}                & \multicolumn{2}{c}{\textbf{LLaMA (33B)}}  & \multicolumn{2}{c}{\textbf{LLaMA (65B)}} \\  
& &{Average} & {Worst}&{Average} & {Worst}&{Average} & {Worst} \\ \midrule\multirow{4}{*}{TREC}  &  Random (cal) &  $66.8_{9.0}$   &  $57.2$   &$69.2_{6.2}$  & $59.4$ &$\pmb{74.6_{9.7}}$  & $\pmb{66.2}$  \\ \cmidrule{2-8} & Ours & $66.8_{2.5}$ & $64.0$ & $\pmb{80.2_{5.3}}$ &$\pmb{75.0}$ & ${74.0_{12.2}}$& ${50.0}$ \\ &Ours (cal)& $\pmb{77.0_{1.1}}$ & $\pmb{75.0}$ & $76.6_{5.1}$ &$70.0$ & $72.8_{12.6 }$ & $48.0$ \\ \midrule \multirow{4}{*}{AGNews}  &   Random (cal) &  $73.0_{6.6}$   & $61.8$   &$71.9_{5.0}$  & $64.0$ &$\pmb{78.2_{4.7}}$   & $\pmb{71.6}$ \\ \cmidrule{2-8} &Ours& $\pmb{79.6_{1.4}}$ &$\pmb{77.0}$  & $\pmb{76.4_{4.6}}$ & $\pmb{69.0}$& $76.0_{4.0}$&$71.0$ \\ &Ours (cal)& $77.4_{1.4}$ & $76.0$ & $76.0_{4.4}$ & $68.0$ & $76.4_{3.6}$ & $70.0$\\  \midrule \multirow{4}{*}{CoLA}  &   Random (cal) &  $\pmb{68.5_{5.5}}$   &$\pmb{61.2}$ &$67.8_{5.1}$  & $63.6$ &$54.0_{12.4}$  & $42.4$  \\ \cmidrule{2-8} &Ours& ${68.2_{6.2}}$ &  $57.0$ & $\pmb{70.6_{4.2}}$ & $64.0$& $\pmb{72.0_{4.5}}$ & $\pmb{66.0}$\\  &Ours (cal)& $68.0_{5.2}$ & ${58.0}$& $70.4_{3.8}$ &$\pmb{65.0}$ & $\pmb{72.0_{4.5}}$ &  $\pmb{66.0}$\\ \bottomrule
\end{tabular}
\end{table}

%% file: secs/appendix.tex
\appendix
\section{Appendix}
\subsection{Pretrained Large Language Models}
Neural autoregressive language model (LMs) are designed for next token prediction to predict the probability distribution over the next token after a sequence of tokens input, and pre-trained LMs show their superior performance since they are trained on various programming languages and a large-scale curated dataset. Training large natural LMs are very expansive and time-consuming process since they always have billions of parameters, which limits the development of LMs. Fortunately, many pre-trained LMs are open access or limited access, which promotes researchers to pool their time and makes the resources to collectively achieve a higher impact. EleutherAI makes the GPT-J~\cite{gpt-j} and GPT-Neox~\cite{gptnexo2022} public available on Hugging Face. GPT-3~\cite{gpt32020brown} is limited access in OpenAI which can be used by researchers for a fee, and another large open-science open-access multilingual language model named Bloom~\cite{bloom2022} is provided by BigScience.


\subsection{Open Access Models}
\label{sec:app-models}
\input{tabs/models}

\subsection{Additional Figures on Different Settings}\label{sec:app-addfigs}
In additional to the Fig.~\ref{fig:allcandidates}, we shows the performance on different models for enumerating all candidates, note that the shadow indicates the half value of standard deviation for clear presentation since the variance is very high for LLMs.

\begin{figure}[ht]
    \centering
    
    \subfloat[AGNews (GPT2-XL 1.5B)]{
    \centering
    \includegraphics[width=0.31\linewidth]{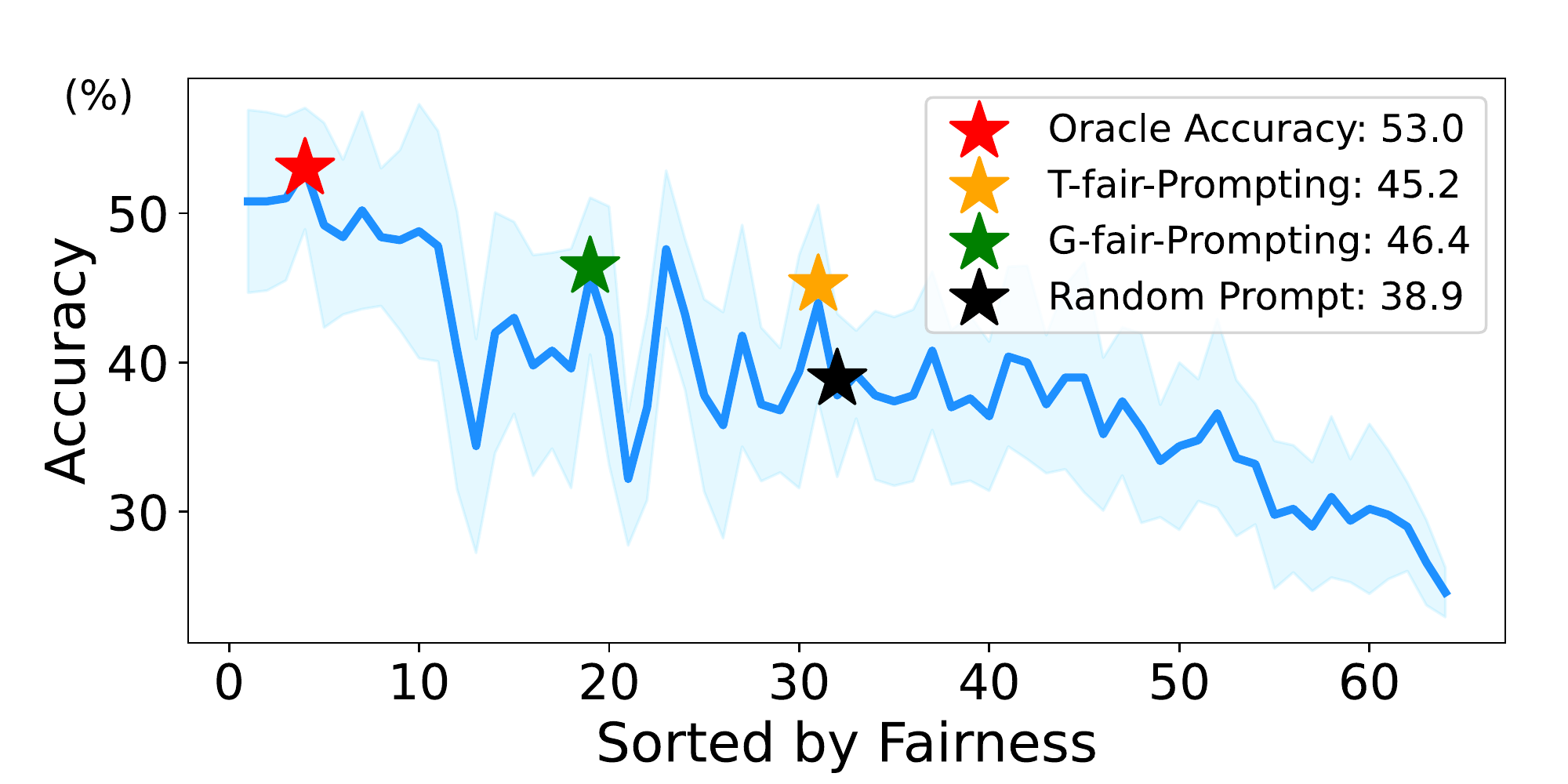}
    } 
    \subfloat[TREC (GPT2-XL 1.5B)]{
    \centering
    \includegraphics[width=0.31\linewidth]{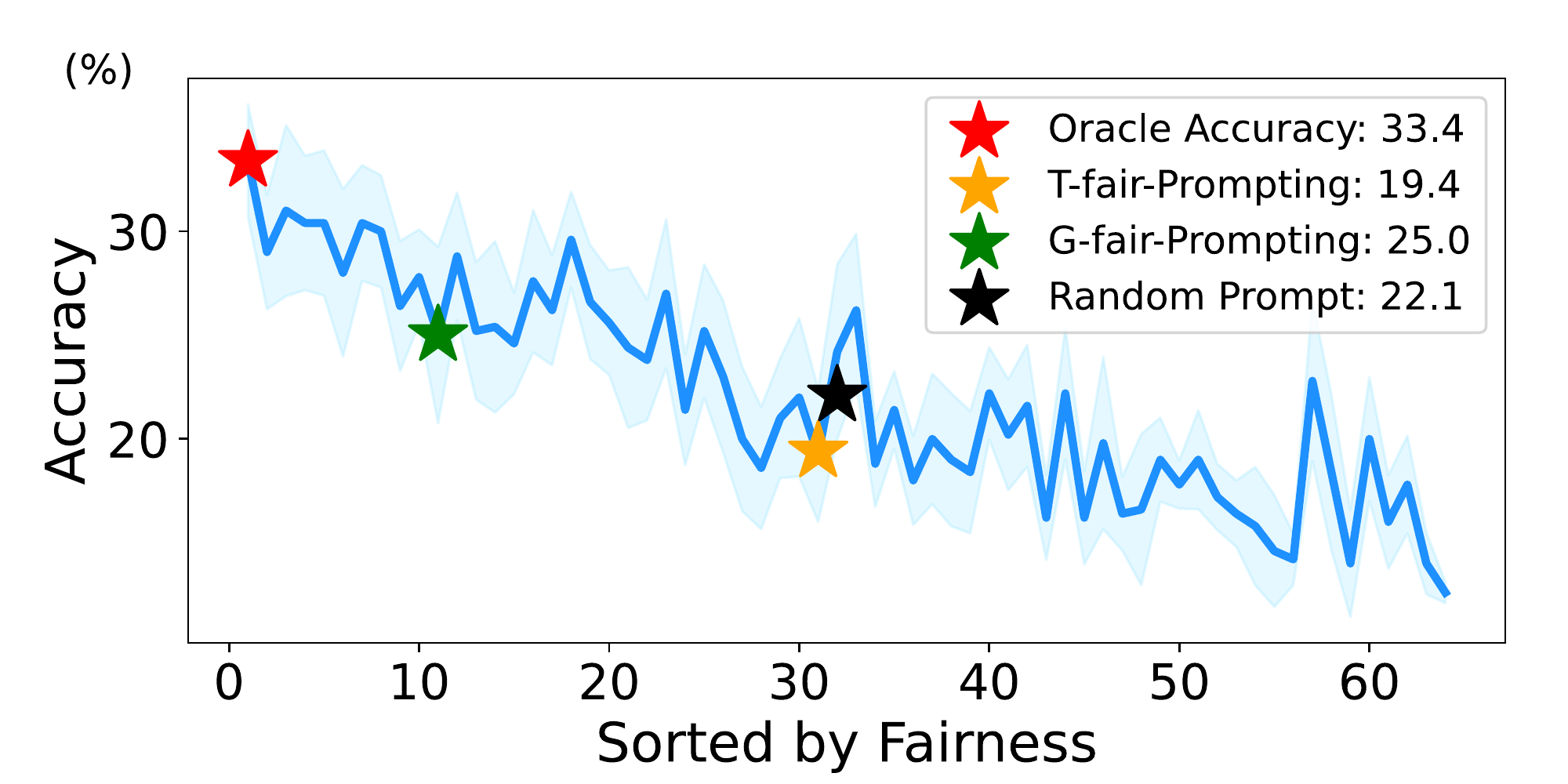}
    } 
    \subfloat[RTE (GPT2-XL 1.5B)]{
    \centering
    \includegraphics[width=0.31\linewidth]{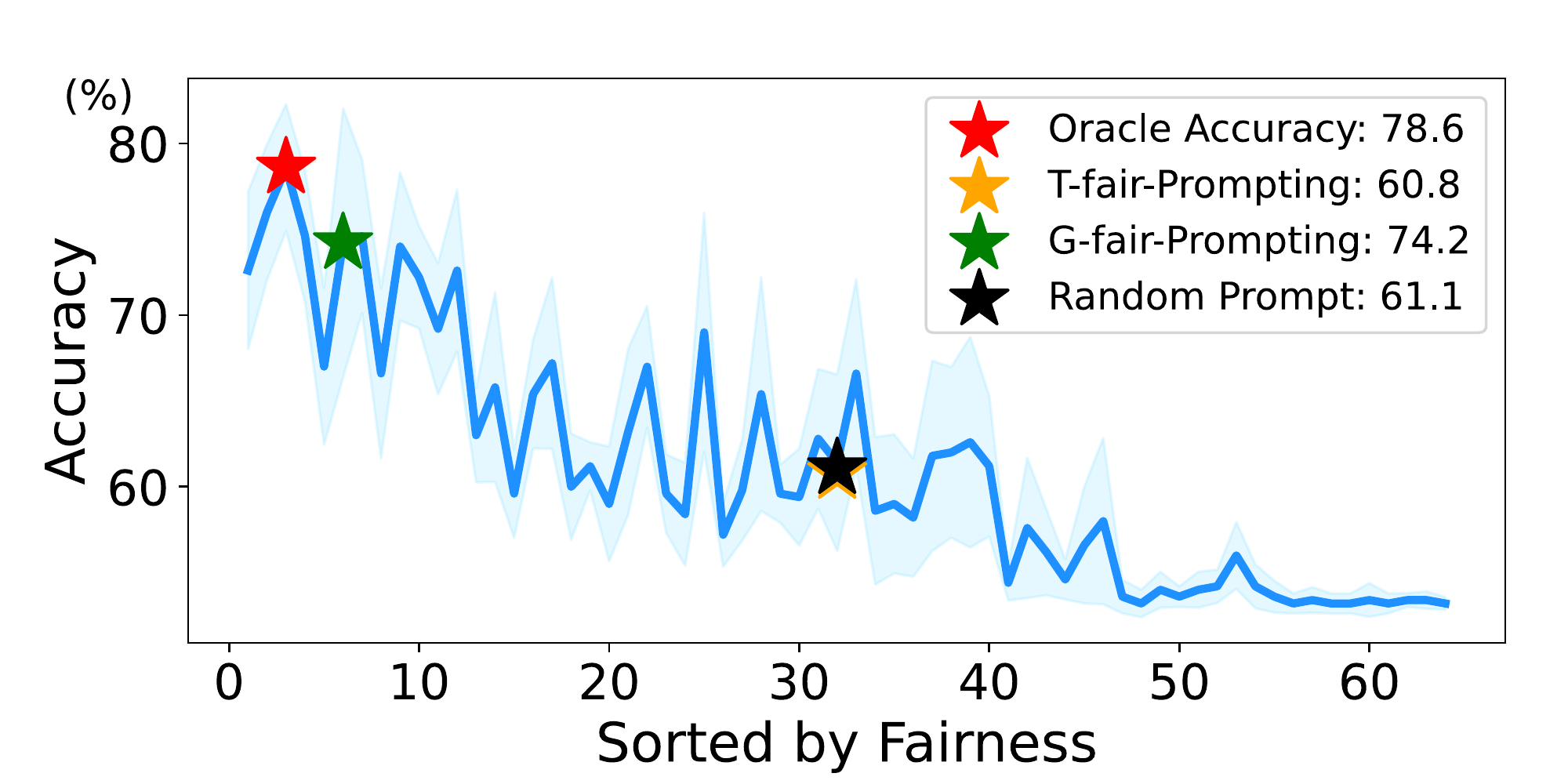}
    }  \\
    \subfloat[AGNews (LLaMA 33B)]{
    \centering
    \includegraphics[width=0.31\linewidth]{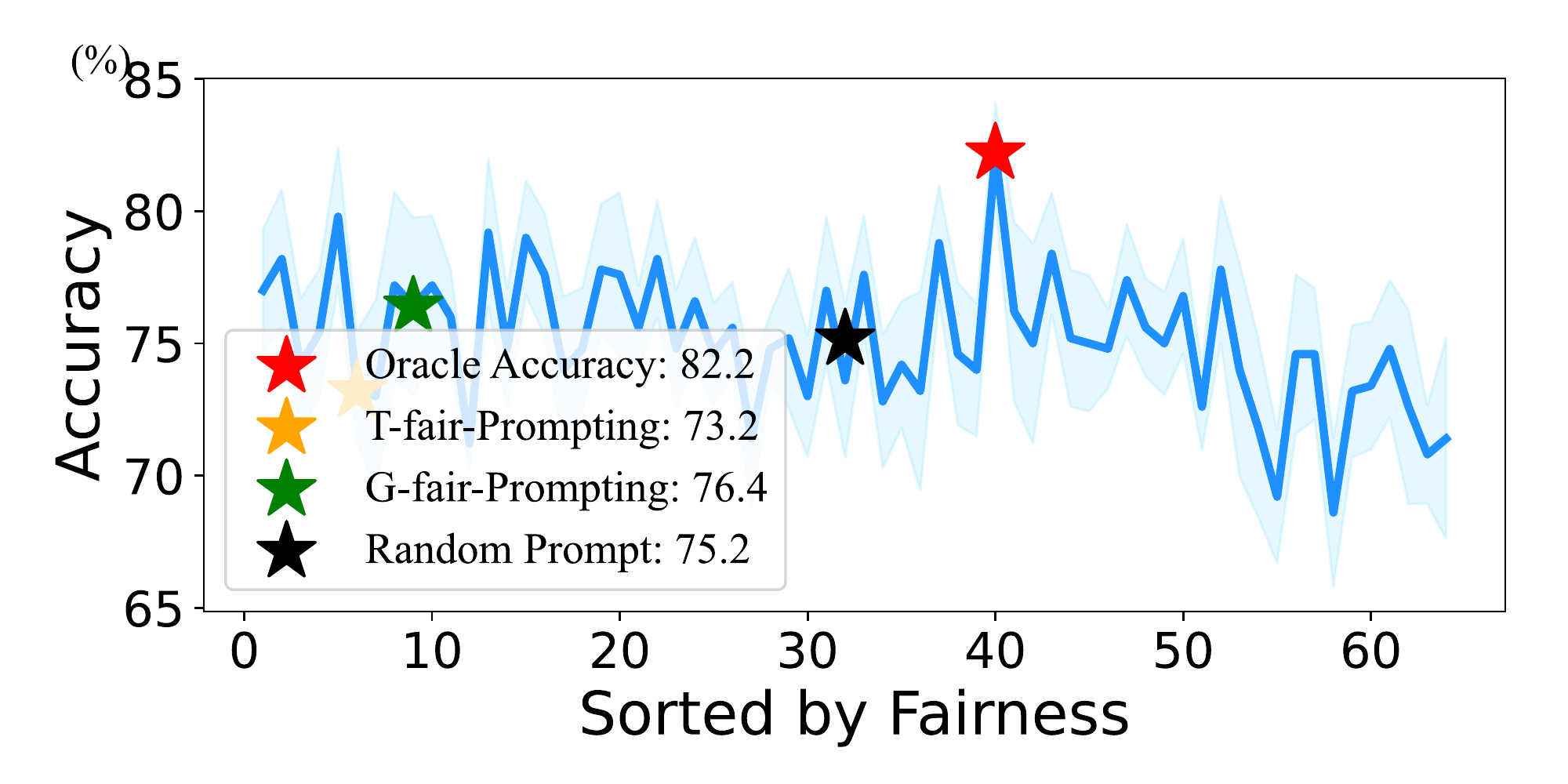}
    }
    \subfloat[TREC (LLaMA 33B)]{
    \centering
    \includegraphics[width=0.31\linewidth]{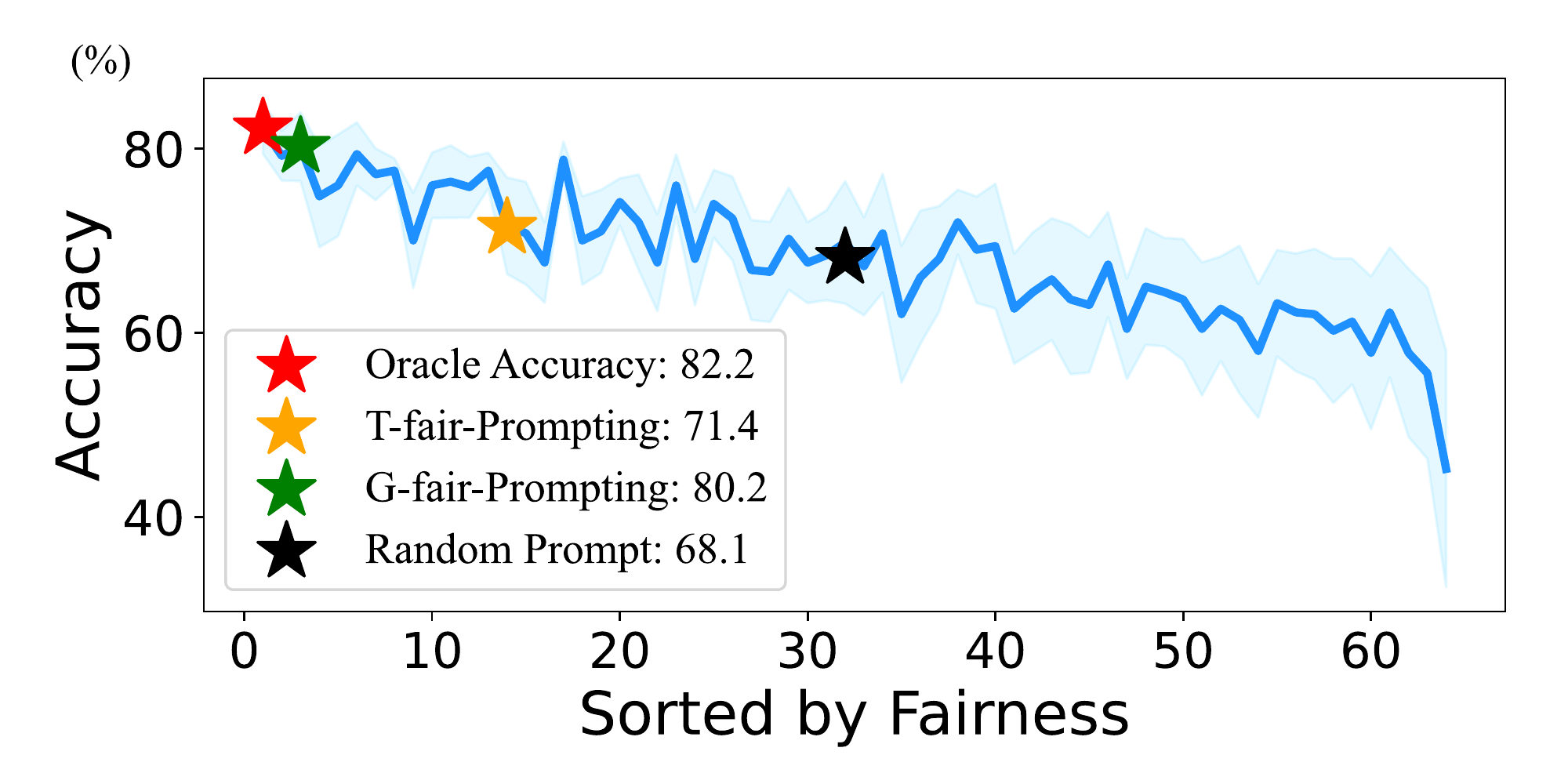}
    }
    \subfloat[SST-2 (LLaMA 33B)]{
    \centering
    \includegraphics[width=0.31\linewidth]{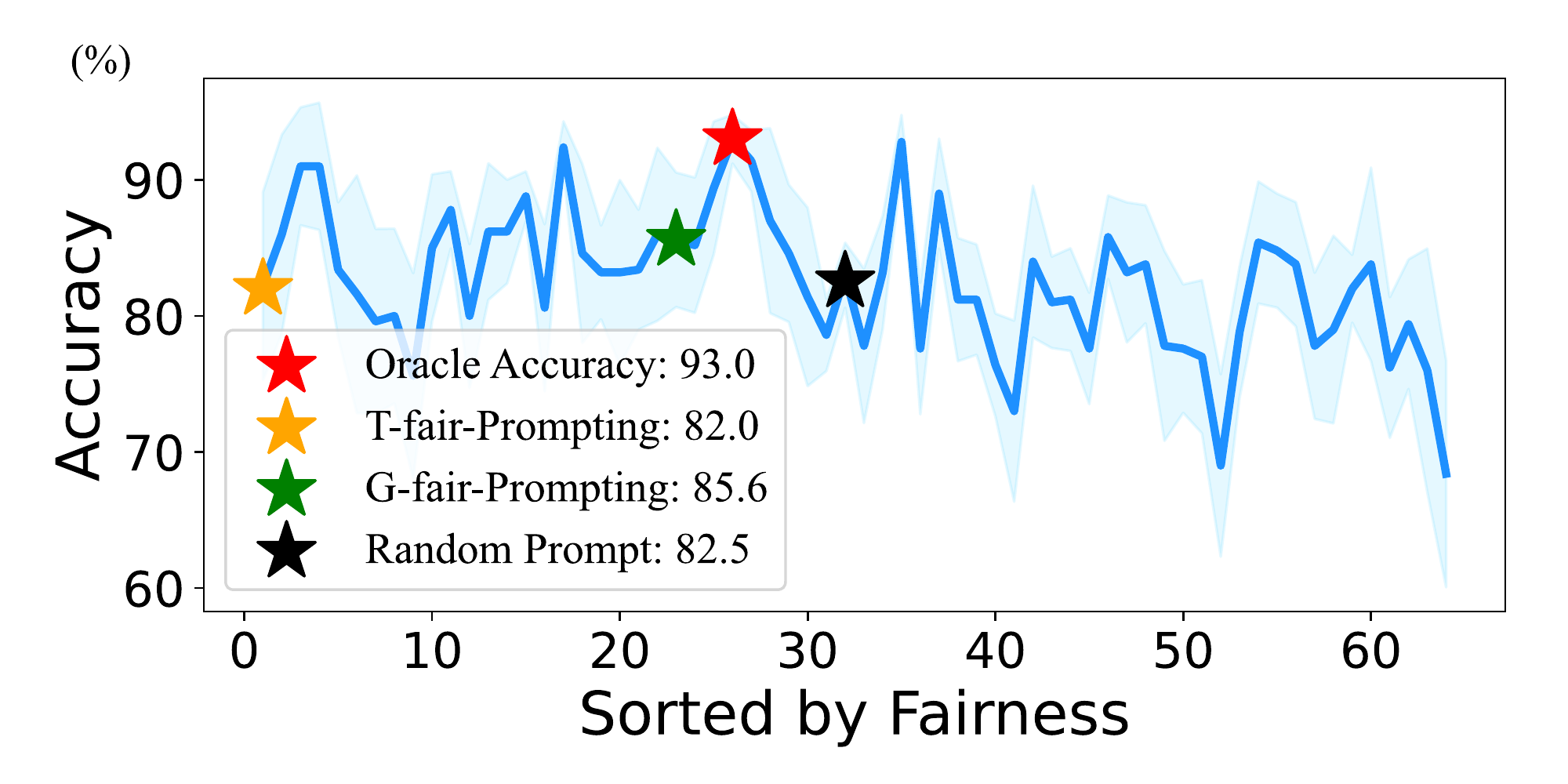}
    } 
    \caption{Accuracy is highly consistency with
fairness and greedy search can find a good prompt, where "Random" and "Oracle" indicates the average accuracy of all prompts and the upper-bound performance according to fairness.}
\label{fig:app-allcandidates}
\end{figure}

\subsection{Accuracy Varies with demonstrations}\label{sec:app-ob}

\begin{figure}[t]
    \centering
    \subfloat[Varying amount of examples]{
    \centering
    \includegraphics[width=0.31\linewidth]{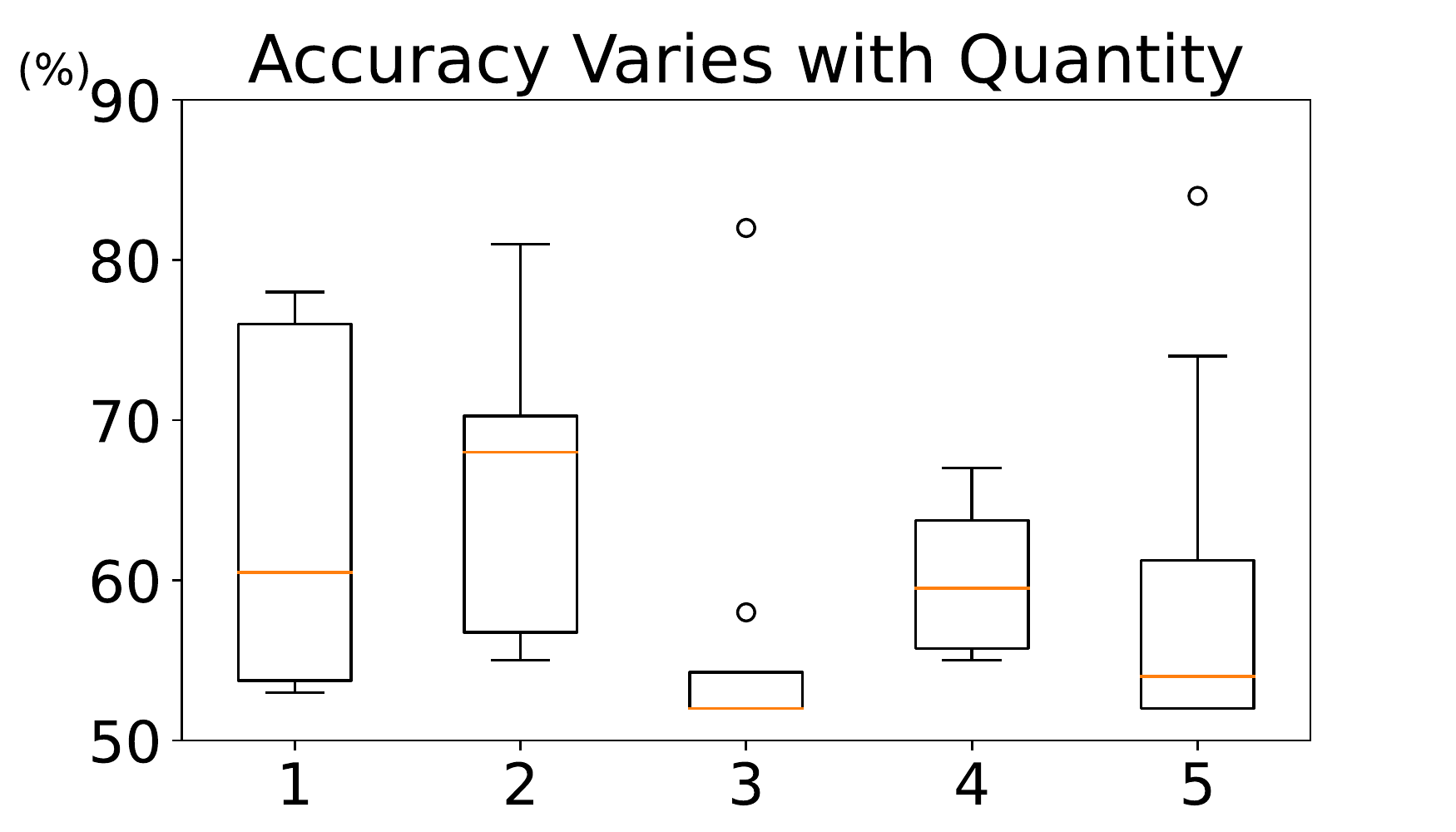}
    \label{fig:sub-quantity}
    }  
    \subfloat[Permutation]{
    \centering
    \includegraphics[width=0.31\linewidth]{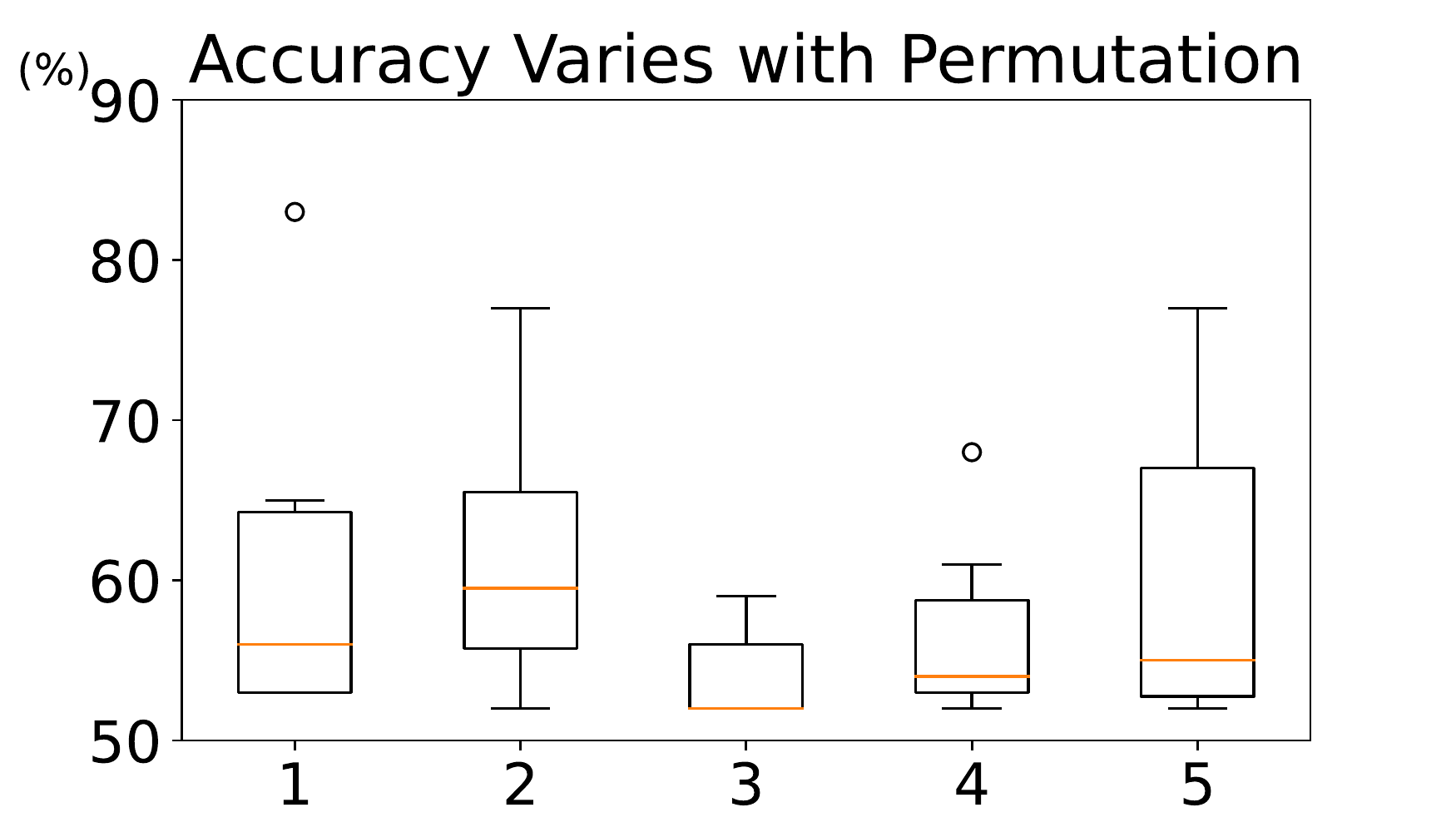}
    \label{fig:sub-order}
    } 
    \subfloat[Select different examples]{
    \centering
    \includegraphics[width=0.31\linewidth]{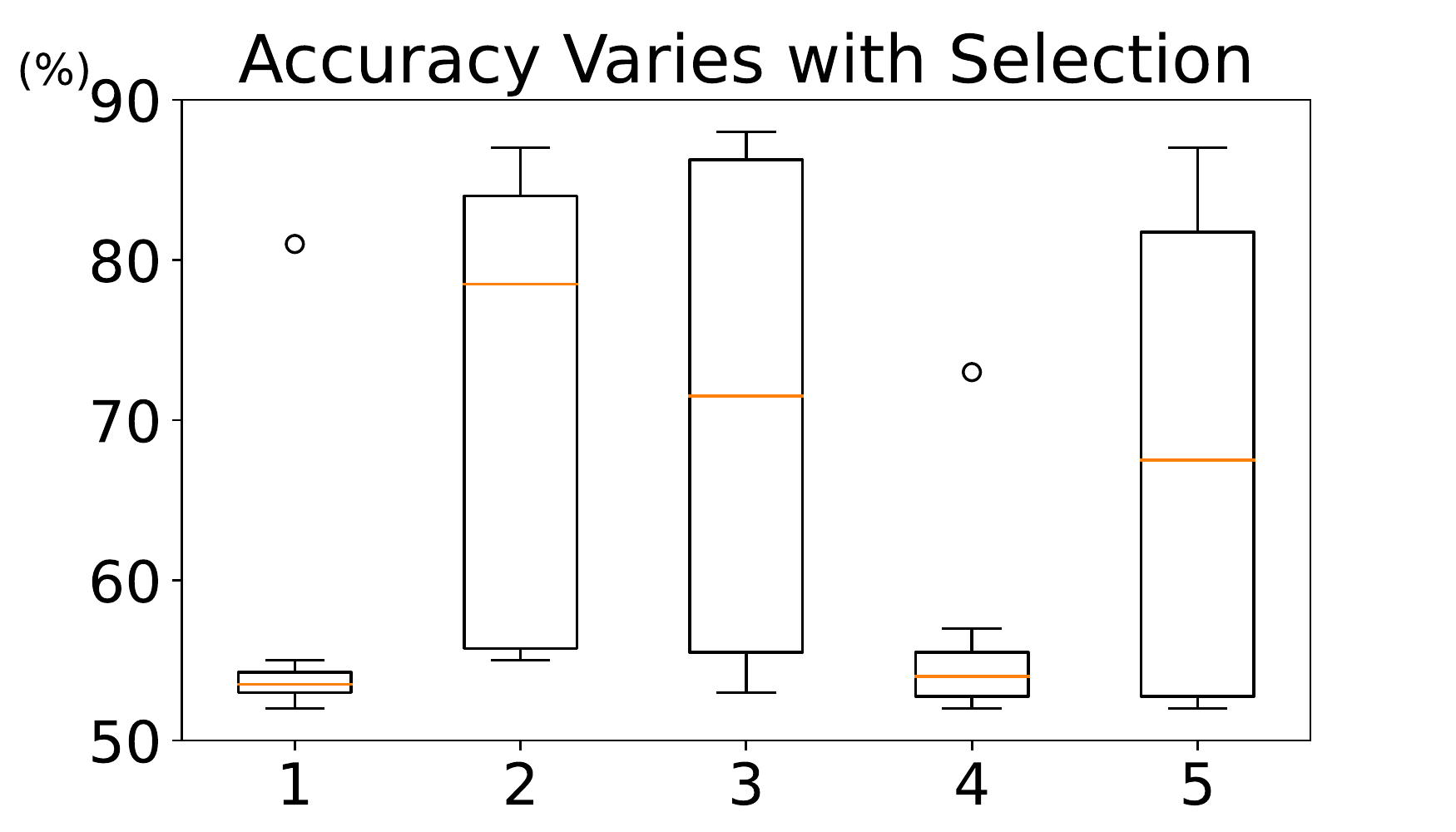}
    \label{fig:sub-select}
    } 
    \caption{ICL suffers from high instability due to variations in example amount, example order, and example selection.}
    \label{fig:app-obser-var}
\end{figure}
\textbf{Accuracy Varies with Example Amount}\quad
Demonstrations play an important role in imparting task-related information to language models through in-context learning. Then, the question arises - does a larger number of demonstrations necessarily equate to better performance? To answer this question, we evaluated performance in terms of accuracy by gradually increasing the number of demonstrations. We set $\rho=\Gamma(x_1,y_1)\oplus\cdots\oplus\Gamma(x_k,y_k)$, where $k =1,\cdots, n$, and demonstrations are erased with $k$ decreasing from $n$ to $1$. Intuitively, accuracy would vary highly across different numbers of demonstrations, and the phenomenon is observed in Fig.~\ref{fig:sub-quantity}. To our surprise, however, erasing some demonstrations can result in a better performance. Removing some demonstrations can perform better and sometimes GPT-3 achieves best accuracy when there is only a few demonstrations remaining. This highlights the importance of considering the appropriate number of demonstrations.

\textbf{Example Order}\quad
The performance of a model is sensitive to the order of the demonstrations, as has been discussed in \cite{order2021lu}. Even when the demonstrations are the same, different permutations of the demonstrations can result in vastly different outcomes. As there are $n!$ possible permutations, we introducing a strategy of permuting the demonstrations by circularly shifting the index of the demonstrations. The demonstration can be represented as $\rho=\Gamma(x_{k+1},y_{k+1})\oplus\cdots\oplus\Gamma(x_n,y_n)\oplus\Gamma(x_1,y_1)\oplus\cdots\oplus\Gamma(x_k,y_k)$.As shown in Fig.~\ref{fig:sub-order}, the accuracy varies highly with permutation which consistent with the observations in \cite{order2021lu}.

\textbf{Example Selection}\quad
In this paper, we find which demonstrations are selected is influence the model extremely. This scenario can be described as selecting $k$ demonstrations in $n$ training samples. In Fig.~\ref{fig:sub-select}, we only select one example for demonstration to ablate the impact of demonstrations order, and the accuracy also varies highly with different example selected. In this work, we only detail evaluate the proposed probing method on the erasing demonstrations and permutation, although our method improves by $20\%$ in the setting of example selection on SST-2 (GPT2-XL), because selecting $k$ demonstrations on a set with $n$ training samples can't be regarded as $k-$shot learning in the strict sense.

\input{tabs/app-accuracy}

\subsection{Relationship between with- and without-calibration}
\begin{figure}[ht]
\centering
  \includegraphics[width=0.75\linewidth]{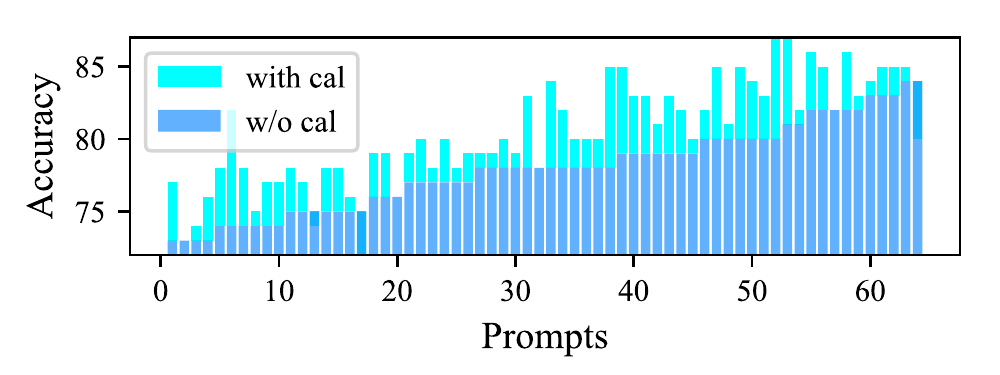}
 \caption{Illustration of accuracy relationship between with- and without calibration when $Pearson$ is positive.}
 \label{fig:app-with-without}
\end{figure}
$\bullet$ \textbf{\greedy without post-calibration outperforms random demonstrations after post-calibration.}
Based on Table~\ref{tab:four-topk-greedy}, it is apparent that \greedy outperforms random selection prior to post-calibration. This leads to a natural question: do prompts with better performance before calibration also indicate better performance after calibration proposed by Zhao et al.~\cite{calibrate2021zhao}? To investigate the relationship between performance with- and without-calibration, we calculated the Pearson correlation coefficient between the accuracy with- and without-calibration $Pearson(acc_{w/o},acc_{with})$. A positive coefficient value suggests that a prompt with high accuracy before calibration has a higher likelihood of achieving higher accuracy after calibration than other prompts. We take the topic classification task on LLaMA(65B) for illustration to show the relationship between with- and without calibration when $Pearson$ is positive in Fig.\ref{fig:app-with-without}. Table~\ref{tab:app-pearson} presents the Pearson correlation coefficient on accuracy of permutation and \greedy after calibration. The majority of Pearson correlation coefficients were found to be positive, indicating that prompts with better performance before calibration have more potential to perform well after calibration. Furthermore, our results on the LLaMA family reveal that the larger the model, the stronger the correlation between performance with- and without-calibration. For instance, the value of the Pearson correlation coefficient increases from $0$ to $0.7$ as the model size increases.

\begin{theorem}
\label{theorem:cal}
Suppose the performance of the model under certain prompts with- and without-calibration is positively correlated, i.e., $Pearson(acc_{w/o},acc_{with})>0$, if we can assure $\mathbb{E}(acc^{Selected}_{w/o})>\mathbb{E}(acc^{Random}_{w/o})$, then we have $\mathbb{E}(acc^{Selected}_{with})>\mathbb{E}(acc^{Random}_{with})$.
\end{theorem}
\input{tabs/app-pearson}

As analysed in Theorem~\ref{theorem:cal}, if we can find a prompt with high accuracy before calibration, we have a higher likelihood of achieving higher accuracy after calibration than random selection. Our approach consistently identifies an appropriate prompt, as evidenced by the results in Table~\ref{tab:four-topk-greedy}. Moreover, the performance of the model exhibits a positive correlation with and without calibration under certain prompts, as illustrated in Table~\ref{tab:app-pearson}. Therefore, our method is more likely to enhance calibration performance.

\subsection{Complexity of different strategies}

\begin{figure}[ht]
\centering
  \includegraphics[width=0.65\linewidth]{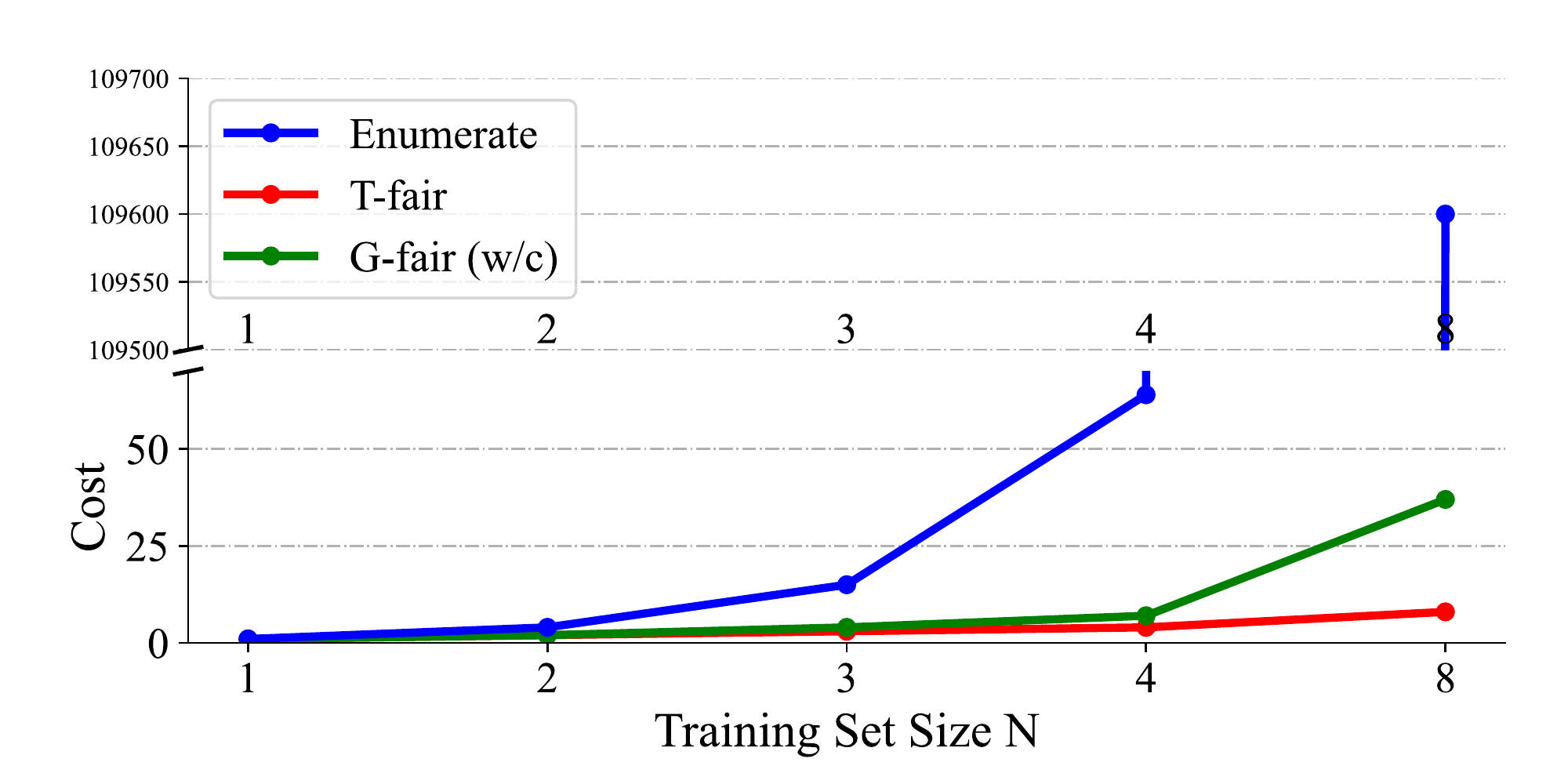}
 \caption{Computational cost. T-fair and G-fair indicate \topk and \greedy respectively, and "w/c" indicates the worst case.}
 \label{fig:app-cost}
\end{figure}

\subsection{Performance on Zero-shot and SOTA Classifiers}

%% file: tabs/models.tex
\begin{table}[ht]
\centering
\caption{Pretrained language models}
\label{tab:models}
\begin{tabular}{lccc}
\toprule
\begin{tabular}[c]{@{}l@{}}\textbf{Model}\\\end{tabular} & \begin{tabular}[c]{@{}l@{}}\textbf{Params}\\\end{tabular} & \begin{tabular}[c]{@{}l@{}}\textbf{Provider}\\\end{tabular} & \begin{tabular}[c]{@{}l@{}}\textbf{Access}\\\end{tabular}  \\
\midrule
GPT-2                                                    & 124 M                                                     & Hugging Face                                                & OPEN                                                       \\
GPT-Medium                                               & 335 M                                                     & Hugging Face                                                & OPEN                                                       \\
GPT2-Large                                               & 774 M                                                     & Hugging Face                                                & OPEN                                                       \\
GPT-XL                                                   & 1.5 B                                                     & Hugging Face                                                & OPEN                                                       \\ \midrule GPT-3 (ada)                                              & 350 M                                                     & OPENAI                                                      & LIMITED                                                    \\
GPT-3 (babbage)                                          & 1.3 B                                                     & OPENAI                                                      & LIMITED                                                    \\
GPT-3 (curie)                                            & 6.7 B                                                     & OPENAI                                                      & LIMITED                                                    \\
GPT-3 (davinci)                                          & 175 B                                                     & OPENAI                                                      & LIMITED   \\  \midrule
GPT-J                                                    & 6 B                                                       & EleutherAI                                                  & OPEN                                                       \\
GPT-NeoX                                                 & 20 B                                                      & EleutherAI                                                  & OPEN                                                       \\ \midrule

Bloom                                                    & 176 B                                                       & BigScience                                                  & OPEN                                                                                                           \\ \midrule
\multirow{4}{*}{LLaMA}                                                   & 7 B                                                       & Meta                                                  & OPEN                                                       \\                      & 13 B                                                       & Meta                                                  & OPEN                                                       \\                     & 33 B                                                       & Meta                                                  & OPEN                                                       \\                     & 65 B                                                       & Meta                                                  & OPEN                                                       \\
\bottomrule
\end{tabular}
\end{table}

%% file: tabs/app-accuracy.tex
\begin{table}[ht]
\centering
\caption{Accuracy for different prompting strategies (averaged on $5_{0,\cdots,4}$ different seeds).}
\label{tab:app-acc}
\resizebox{1.0\textwidth}{!}{\begin{tabular}{c|c|ccc||ccc} \toprule
\multirow{2}{*}{\textbf{Model}}                  & \multirow{2}{*}{\textbf{Dataset}}                  & \multirow{2}{*}{\textbf{Random}} & 
 \multirow{2}{*}{\textbf{Diversity}} & \multirow{2}{*}{\textbf{Similarity}} & \multicolumn{3}{c}{\textbf{Ours}} \\ & & & &  &\textbf{Top-2} & \textbf{Top-4} & \textbf{Greedy}                \\ \midrule
\multirow{5}{*}{GPT2-XL (1.5B)} & {SST-2} & $61.1_{6.1}$  &$-$      & $-$   &$60.8_{11.4}$      & $65.8_{8.7}$       & $74.2_{12.0}$    \\  \cmidrule{2-8} & {AGNews} & $38.9_{11.4}$  &$-$      & $-$   &$45.2_{12.5}$      & $37.2_{11.2}$       & $46.4_{11.9}$    \\  \cmidrule{2-8}
& {TREC} & $22.1_{5.7}$  &$-$      & $-$   &$19.4_{8.9}$      & $28.2_{9.2}$       & $25.0_{7.4}$    \\  \cmidrule{2-8}
& {RTE} & $53.2_{6.9}$  &$-$      & $-$   &$54.0_{7.5}$      & $53.6_{5.9}$       & $56.4_{2.2}$   
\\ \midrule \multirow{4}{*}{LLaMA (7B)}  &    {AGNews}                             & $64.5_{10.0}$  &${66.4_{9.1}}$      & $-$   &$66.0_{11.7}$      & ${69.2_{5.5}}$       & $63.8_{5.7}$    \\  \cmidrule{2-8}
                                & {TREC}  & $49.5_{10.4}$ &${51.4_{9.6}}$            &  {$-$}       &$48.4_{10.5}$            & $38.6_{15.2}$      & ${61.3_{4.8}}$   \\ \cmidrule{2-8}
                                & {CoLA} & $60.4_{10.6}$ &$63.8_{8.7}$& $-$  &$58.2_{7.8}$            & $61.6_{6.5}$       & $36.4_{3.6}$ \\ \midrule\multirow{4}{*}{LLaMA (13B)}  &   {AGNews}                         & $72.2_{7.7}$ &$78.4_{3.5}$      & {$-$} &$73.6_{9.0}$      & $74.2_{4.3}$            & ${75.2_{2.8}}$    \\ \cmidrule{2-8}
                                & {TREC}  & $46.4_{16.5}$ &$48.0_{16.0}$            & $-$   &$51.0_{16.6}$            & $39.2_{23.3}$        & ${61.4_{12.1}}$  \\ \cmidrule{2-8}
                                & {CoLA} & $67.7_{2.9}$ &$67.2_{2.4}$& $-$  &$67.0_{2.0}$            & $67.2_{1.6}$       & ${67.0_{2.0}}$ \\  \bottomrule
\end{tabular}}
\end{table}

%% file: tabs/app-pearson.tex
\begin{table}[ht]
\centering
\caption{Pearson's r between the with- and without-calibration.}
\label{tab:app-pearson}
\begin{tabular}{c||ccccc} \toprule
\multirow{2}{*}{\textbf{Dataset}}                    & {\textbf{BLOOM}}                & \multicolumn{4}{c}{\textbf{LLaMA}} \\  & \textbf{176B} &\textbf{7B} & \textbf{13B} & \textbf{33B}  & \textbf{65B}                 \\ \midrule{TREC}    &  $0.1274$   & $0.1551$ &  $0.2959$   &$0.3090$    &$0.5151$     \\ {AGNews}  &  $0.3875$   & $-0.0471$ &  $0.3044$   &$0.6953$    &$0.7100$   \\  {CoLA} &  $0.4050$   & $0.3592$ &  $0.5193$   &$0.3611$    &$0.8012$   \\ \bottomrule
\end{tabular}
\end{table}

%% file: main.bbl
\begin{thebibliography}{10}

\bibitem{gpt32020brown}
Tom Brown, Benjamin Mann, Nick Ryder, Melanie Subbiah, Jared~D Kaplan, Prafulla
  Dhariwal, Arvind Neelakantan, Pranav Shyam, Girish Sastry, Amanda Askell,
  Sandhini Agarwal, Ariel Herbert-Voss, Gretchen Krueger, Tom Henighan, Rewon
  Child, Aditya Ramesh, Daniel Ziegler, Jeffrey Wu, Clemens Winter, Chris
  Hesse, Mark Chen, Eric Sigler, Mateusz Litwin, Scott Gray, Benjamin Chess,
  Jack Clark, Christopher Berner, Sam McCandlish, Alec Radford, Ilya Sutskever,
  and Dario Amodei.
\newblock Language models are few-shot learners.
\newblock In {\em NeurIPS}, volume~33, pages 1877--1901, 2020.

\bibitem{bloom2022}
Bloom: A 176b-parameter open-access multilingual language model.
\newblock \url{https://huggingface.co/bigscience/bloom}.

\bibitem{petroni2019language}
Fabio Petroni, Tim Rockt{\"a}schel, Patrick Lewis, Anton Bakhtin, Yuxiang Wu,
  Alexander~H Miller, and Sebastian Riedel.
\newblock Language models as knowledge bases?
\newblock {\em arXiv preprint arXiv:1909.01066}, 2019.

\bibitem{order2021lu}
Yao Lu, Max Bartolo, Alastair Moore, Sebastian Riedel, and Pontus Stenetorp.
\newblock Fantastically ordered prompts and where to find them: Overcoming
  few-shot prompt order sensitivity.
\newblock In {\em ACL}, 2021.

\bibitem{nie2022improving}
Feng Nie, Meixi Chen, Zhirui Zhang, and Xu~Cheng.
\newblock Improving few-shot performance of language models via nearest
  neighbor calibration.
\newblock {\em arXiv preprint arXiv:2212.02216}, 2022.

\bibitem{liu2023pre}
Pengfei Liu, Weizhe Yuan, Jinlan Fu, Zhengbao Jiang, Hiroaki Hayashi, and
  Graham Neubig.
\newblock Pre-train, prompt, and predict: A systematic survey of prompting
  methods in natural language processing.
\newblock {\em ACM Computing Surveys}, 55(9):1--35, 2023.

\bibitem{li2021prefix}
Xiang~Lisa Li and Percy Liang.
\newblock Prefix-tuning: Optimizing continuous prompts for generation, 2021.

\bibitem{liu2021p}
Xiao Liu, Kaixuan Ji, Yicheng Fu, Zhengxiao Du, Zhilin Yang, and Jie Tang.
\newblock P-tuning v2: Prompt tuning can be comparable to fine-tuning
  universally across scales and tasks.
\newblock {\em arXiv preprint arXiv:2110.07602}, 2021.

\bibitem{hambardzumyan2021warp}
Karen Hambardzumyan, Hrant Khachatrian, and Jonathan May.
\newblock Warp: Word-level adversarial reprogramming.
\newblock {\em arXiv preprint arXiv:2101.00121}, 2021.

\bibitem{qin2021learning}
Guanghui Qin and Jason Eisner.
\newblock Learning how to ask: Querying lms with mixtures of soft prompts.
\newblock {\em arXiv preprint arXiv:2104.06599}, 2021.

\bibitem{liu2021gpt}
Xiao Liu, Yanan Zheng, Zhengxiao Du, Ming Ding, Yujie Qian, Zhilin Yang, and
  Jie Tang.
\newblock Gpt understands, too.
\newblock {\em arXiv preprint arXiv:2103.10385}, 2021.

\bibitem{zhang2022automatic}
Zhuosheng Zhang, Aston Zhang, Mu~Li, and Alex Smola.
\newblock Automatic chain of thought prompting in large language models.
\newblock {\em arXiv preprint arXiv:2210.03493}, 2022.

\bibitem{gentile2022fast}
Claudio Gentile, Zhilei Wang, and Tong Zhang.
\newblock Fast rates in pool-based batch active learning.
\newblock {\em arXiv preprint arXiv:2202.05448}, 2022.

\bibitem{diao2023active}
Shizhe Diao, Pengcheng Wang, Yong Lin, and Tong Zhang.
\newblock Active prompting with chain-of-thought for large language models.
\newblock {\em arXiv preprint arXiv:2302.12246}, 2023.

\bibitem{liu2021makes}
Jiachang Liu, Dinghan Shen, Yizhe Zhang, Bill Dolan, Lawrence Carin, and Weizhu
  Chen.
\newblock What makes good in-context examples for gpt-$3 $?
\newblock {\em arXiv preprint arXiv:2101.06804}, 2021.

\bibitem{shi2023large}
Freda Shi, Xinyun Chen, Kanishka Misra, Nathan Scales, David Dohan, Ed~Chi,
  Nathanael Sch{\"a}rli, and Denny Zhou.
\newblock Large language models can be easily distracted by irrelevant context.
\newblock {\em arXiv preprint arXiv:2302.00093}, 2023.

\bibitem{chatgpt}
\url{https://openai.com/blog/chatgpt}.

\bibitem{calibrate2021zhao}
Tony~Z. Zhao, Eric Wallace, Shi Feng, Dan Klein, and Sameer Singh.
\newblock Calibrate before use: Improving few-shot performance of language
  models.
\newblock In {\em ICML}, 2021.

\bibitem{touvron2023llama}
Hugo Touvron, Thibaut Lavril, Gautier Izacard, Xavier Martinet, Marie-Anne
  Lachaux, Timoth{\'e}e Lacroix, Baptiste Rozi{\`e}re, Naman Goyal, Eric
  Hambro, Faisal Azhar, Aurelien Rodriguez, Armand Joulin, Edouard Grave, and
  Guillaume Lample.
\newblock Llama: Open and efficient foundation language models.
\newblock {\em arXiv preprint arXiv:2302.13971}, 2023.

\bibitem{gpt22018}
Alec Radford, Jeffrey Wu, Rewon Child, David Luan, Dario Amodei, and Ilya
  Sutskever.
\newblock Language models are unsupervised multitask learners.
\newblock {\em Technical Report}, 2018.

\bibitem{wang2018glue}
Alex Wang, Amanpreet Singh, Julian Michael, Felix Hill, Omer Levy, and Samuel~R
  Bowman.
\newblock Glue: A multi-task benchmark and analysis platform for natural
  language understanding.
\newblock {\em arXiv preprint arXiv:1804.07461}, 2018.

\bibitem{gpt-j}
Ben Wang and Aran Komatsuzaki.
\newblock {GPT-J-6B: A 6 Billion Parameter Autoregressive Language Model}.
\newblock \url{https://github.com/kingoflolz/mesh-transformer-jax}, May 2021.

\bibitem{gptnexo2022}
\url{https://huggingface.co/EleutherAI/gpt-neox-20b}.

\end{thebibliography}
